\documentclass[lettersize,journal]{IEEEtran}
\usepackage{circledsteps}
\usepackage{amsmath,amsfonts}
\usepackage{array}
\usepackage{textcomp}
\usepackage{stfloats}
\usepackage{url}
\usepackage{verbatim}
\usepackage{graphicx}
\usepackage{cite}
\usepackage{cuted}
\usepackage{hyperref}
\usepackage{amsthm}
\usepackage{subcaption}
\usepackage{soul}
\usepackage{xcolor}
\usepackage[linesnumbered,ruled,vlined]{algorithm2e}
\usepackage{pifont}
\usepackage{anyfontsize}

\usepackage[switch]{lineno}
\usepackage{multirow}
\usepackage{balance}
\usepackage{booktabs}

\begin{document}

\title{Cross-Area Transfer Learning for VLC-based Indoor Localization with a Transfer Efficiency Score  }

\author{Masood Jan, Wafa Njima, Xun Zhang, Alexander Artemenko

\thanks{Masood Jan is with the Doctoral School of Computer Science, Telecommunications and Electronics (EDITE), Sorbonne University, 75006 Paris, France and affiliated with the LISITE Laboratory, Institut Supérieur d'Électronique de Paris, 75006 Paris, France.(e-mail: masoodjann@gmail.com). \textit{(Corresponding authors Masood Jan, Xun Zhang.)} }
\thanks{Wafa Njima is with the LISITE Laboratory, Institut Supérieur d'Électronique de Paris, France. (e-mail: wafa.njima@isep.fr).}
\thanks{Xun Zhang is with the LISITE Laboratory, Institut Supérieur d'Électronique de Paris and also affiliated with the Université Paris-Saclay, LISV Laboratory, France, (e-mail: xun.zhang@isep.fr).}
\thanks{ Alexander Artemenko is with the sector research of Robert Bosch GmbH in Stuttgart, Germany.  (e-mail: alexander.artemenko@de.bosch.com).}

\thanks{\textcopyright{} 2026 IEEE. Personal use of this material is permitted. Permission from IEEE must be obtained for all other uses, in any current or future media, including reprinting/republishing this material for advertising or promotional purposes, creating new collective works, for resale or redistribution to servers or lists, or reuse of any copyrighted component of this work in other works.}

\thanks{Published in IEEE Internet of Things Journal, vol. 13, no. 15, pp. 35237--35256, 2026. DOI: \url{https://doi.org/10.1109/JIOT.2026.3697560}.}
}

\maketitle

\begin{abstract}
Indoor localization systems are critical for Industry-4.0 applications.
Visible Light Communication (VLC)-based systems offer advantages such as immunity to electromagnetic interference, high accuracy, and energy efficiency, but their performance is hindered by environmental variability across deployment areas. This paper presents a transfer learning (TL) framework for VLC-based indoor localization that enhances cross-area robustness while reducing data requirements. Using real-world measurements from four distinct areas of a Bosch manufacturing facility, we show that fine-tuning a deep neural network (DNN) pre-trained on source area data with only 40\% of target-area measurements recovers 97\% of full-data performance, thereby reducing the amount of target data needed. We further introduce the Transfer Efficiency Score (TES), a composite metric that identifies the most effective source models for transfer without exhaustive evaluation. Guided by TES, the framework selects the best source model, yielding localization errors as low as 40.9 cm with success rates above 86\%. Validation on an unseen area confirms its generalization capability, with a 3.28 cm mean error achieved using the TES-selected source model. Overall, the proposed framework provides a scalable and data-efficient pathway for industrial VLC localization, substantially reducing deployment costs and computational burden while maintaining high accuracy in diverse and dynamic environments.
\end{abstract}

\begin{IEEEkeywords}
Deep Neural Network (DNN), Fine-tuning, Indoor localization, Transfer Learning (TL), Transfer Efficiency Score (TES), Visible Light Communication (VLC).
\end{IEEEkeywords}
\section{Introduction}
\IEEEPARstart{U}{nplanned} downtime from misplaced or untraceable assets remains a major contributor to production delays and operational inefficiencies, underscoring the critical need for robust indoor-positioning solutions in Industry 4.0 environments \cite{yang2024positioning, franko2020reliable}. Modern factories are increasingly transformed by AI-driven automation and digital-twin workflows, with real-time location-based services (LBS), power asset tracking, process optimization, and coordinated robotics \cite{marasovic2024indoor, ullah2025digital}. In this context, indoor localization has evolved from an auxiliary feature into a critical infrastructure that underpins next-generation industrial automation \cite{aziz2024enhancing}.

Despite advances, three core challenges prevent large-scale deployment of indoor localization. First, cross-area robustness, where models trained in one area often degrade significantly when deployed in a different layout or building area \cite{pan2008transfer}. Second, data inefficiency, which means collecting and labeling comprehensive datasets can require days of downtime per area, contradicting the high-cycle production demands \cite{loffredo2023reinforcement, wu2017automatic}. Third, energy and cost constraints, as many radio frequency (RF) based systems consume excessive power or rely on dense beacon infrastructures, driving up maintenance costs and reducing battery lifespan in IoT deployments \cite{sherazi2020energy, milano2024ble}.

While several indoor localization technologies are widely used in industrial environments, each presents its own set of trade-offs relative to the core challenges. Wi‑Fi localization offers moderate power consumption \cite{dai2023survey} but suffers severe accuracy drops under electromagnetic interference (EMI) from metal machinery and dynamic layouts, which affects both robustness and maintenance demands \cite{gonzalez2019energy}. Bluetooth Low Energy (BLE) reduces power consumption in sleep-mode to as little as 10.1 µA \cite{lindemann2016indoor} but is limited by short range and environmental variability, making it costly to scale and sensitive to changes across deployment areas \cite{milano2024ble, wang2025ble}. Both BLE and Wi-Fi solutions depend on dense infrastructure, often requiring a large number of access points or beacons, which introduces high installation and ongoing maintenance costs \cite{guidara2019energy, xiao2016survey}. Ultra‑Wideband (UWB) systems achieve higher accuracy, but come with higher hardware costs and power requirements. Like all RF-based approaches, UWB also remains vulnerable to signal degradation from EMI and dynamic industrial layouts \cite{khriji2022design, schmidt2021detection, aziz2025comprehensive}. These trade-offs highlight the need for a localization approach that blends high accuracy with an efficient and minimal infrastructure.

Visible Light Communication (VLC) localization emerges as a compelling alternative that leverages light-emitting diode (LED) lighting infrastructures to enable communication and positioning. Compared to RF technologies, VLC achieves centimeter-level accuracy and up to 60\% lower energy consumption, while being inherently immune to EMI \cite{almadani2020visible, miranda2023review, wang2024application}. Because VLC relies on unobstructed optical paths, its line‑of‑sight, short‑range nature reduces multipath effects and simplifies the signal‑processing pipeline \cite{ma2023centimeter}. At the same time, this reliance on direct optical paths makes VLC systems sensitive to changes in ambient lighting, temporary blockages, and uneven surface reflections commonly found in industrial environments \cite{cuailean2024driving, apolo2025experimental, zhao2024ultra}. Although these optical challenges can affect performance, they follow consistent physical laws and environmental patterns. For example, ambient lighting often changes gradually or follows known schedules (such as day/night cycles or factory lighting controls); occlusions occur when objects block the line of sight, and their effects can be estimated using geometry or motion sensors; and surface reflectance depends on material properties (e.g., metal or plastic), which influence light reflection in predictable ways. This predictability allows them to be effectively addressed through model-based correction and area-specific adaptation \cite{oluwole2024exploration, santhanalakshmivlc}. 

The complex, non‑linear distortions introduced by EMI, multipath reflections, and dynamic factory layouts have motivated a shift toward machine‑learning (ML) methods, which can learn mapping functions between raw sensor inputs and positions without requiring manual feature engineering or signal modeling \cite{ghonimindoor2024}. However, standard ML paradigms assume that the deployment environment remains static, leading to significant degradation in localization accuracy and generalizability when floorplans, material composition, or illumination conditions change \cite{pan2008transfer}. 

Several adaptation strategies have been explored to address variability across deployment areas. Federated Learning (FL) preserves privacy by distributing model training across edge devices \cite{jan2024privacy}, but it requires periodic exchange of gradient updates over potentially unreliable factory wireless links, introducing latency in highly dynamic or resource‑constrained settings. Continuous Learning (CL) enables incremental adaptation \cite{wickramasinghe2023continual}, yet suffers from catastrophic forgetting and depends on continual access to new labeled data, which is costly and labor-intensive. Transfer Learning (TL), by contrast, requires periodic fine-tuning on minimal datasets from the target deployment area as environmental conditions evolve, thus reducing both communication overhead and the amount of target data required while maintaining high accuracy compared to continuous retraining approaches \cite{su2023four, zhu2024survey, wang2024learning}. 

Despite its success in RF-based localization, TL has seen limited application in VLC systems because fundamental propagation differences create unique adaptation challenges. Unlike RF, which is dominated by EMI and multipath fading, VLC is governed by illumination geometry and line-of-sight dependence. VLC systems therefore face environmental factors such as ambient-light interference (e.g., from fluorescent lamps or welding), geometry-dependent multipath scattering from reflective surfaces, and occlusion patterns that critically affect signal visibility. These VLC-specific characteristics require TL strategies that account for optical propagation properties, ambient lighting conditions, and the inherent directional nature of VLC.

Given these advantages, VLC stands out as a strong candidate for scalable, high-precision indoor localization. However, to fully unlock its potential particularly in the context of cross-area deployments, there is a pressing need for ML solutions that are not only energy-efficient and accurate, but also capable of adapting to area-specific variations in optical conditions. Equally important is the need for systematic model selection mechanisms that can identify the most suitable pre-trained model for a given target deployment without requiring exhaustive fine-tuning of all available candidates. In practical industrial deployments, engineers often have access to multiple pre-trained models from different source areas, each with varying degrees of compatibility with the target environment. Without reliable transferability assessment metrics, practitioners must resort to costly trial-and-error approaches that contradict the efficiency goals of TL. A normalized composite metric that combines multiple performance indicators would enable rapid, cost-effective identification of optimal source models while providing consistent evaluation across diverse deployment scenarios. These considerations motivate our development of a TL-based VLC localization framework that maximizes cross-area adaptability and precision, addressing the interplay between optical variability and ML model generalization.
Our key contributions are as follows:

\begin{itemize}
  \item \textbf{TL Framework for VLC Cross-Area Localization:} We develop and validate a comprehensive TL framework that enables VLC localization models trained in one area to adapt effectively to new layouts with minimal retraining, achieving robust cross-area deployment in dynamic industrial environments. This work is intended as a proof-of-concept demonstrating the suitability and practicality of TL for VLC-based indoor localization under realistic industrial deployment conditions.
  \item \textbf{Data-Efficient Fine-Tuning Strategy:} We demonstrate that fine-tuning on just 40\% of target-area data recovers 97\% of full-data performance, reducing both target-area data collection requirements and deployment effort by 60\% compared to traditional full-retraining approaches.
  \item \textbf{Transfer Efficiency Score (TES) Metric and Validation:} We introduce TES, a normalized composite metric that combines localization error and success rate to systematically select optimal source models without exhaustive testing, and validate its effectiveness through superior performance demonstration on an unseen facility layout.
\end{itemize}

\noindent The remainder of this paper is organized as follows: Section \ref{sec:relworks} reviews related work, Section \ref{sec:sysm} presents the system model and problem formulation, Section~\ref{sec:algorithm} details the proposed algorithm development, Section \ref{sec:setup} discusses the experimental setup and data collection process, Section \ref{sec:results} analyzes performance evaluation results, and Section \ref{sec:conclusion} concludes the paper.

\section{Related Works}
\label{sec:relworks}

The rapid convergence of Industry 4.0 requirements, and advancements in ML techniques has led to extensive research into indoor localization. In particular, cross‐area robustness, data and energy efficiency, and scalable deployment have become focal points. Cross-area robustness refers to the model-level challenge of maintaining localization accuracy when deployed across different physical environments with varying layouts, structural characteristics, and interference patterns, which we address through TL. Scalable deployment, in contrast, represents the system-level challenge of deploying localization across multiple areas and devices without proportional increases in retraining, data collection, or infrastructure costs — a goal facilitated by our data-efficient fine-tuning and TES-based source model selection. We organize this section around three major ML paradigms enabling adaptive localization systems; FL, CL, and TL. We first assess their application in RF-based systems, then examine VLC-based approaches, and finally highlight the bottlenecks that motivate our TL-driven VLC framework.

\subsection{FL for Indoor Localization}
FL has emerged as a promising solution for indoor localization. It enables collaborative model training across edge devices without transferring raw data, thereby reducing communication overhead and preserving user privacy \cite{zhang2020transfer, nguyen2021federated}. Several studies have explored FL in this context. FeMLoc employs a two-stage meta-training and fine-tuning strategy to capture environmental variability, achieving up to 80.9\% gains in accuracy and an 82.2\% reduction in training iterations compared to centralized approaches \cite{etiabi2024femloc}. \cite{kumar2023confidentiality} present a FL framework for Wi-Fi fingerprinting that maintains data privacy on edge devices, enhances localization accuracy, and lowers energy consumption by leveraging distributed computing. \cite{jan2024privacy} demonstrate that retaining all training data locally and transmitting only model updates such as gradients or weights can significantly reduce bandwidth usage while ensuring sensitive sensor data remains confined to local hardware. Similarly, \cite{etiabi2023federated} further enhance generalization by randomizing and distributing building-specific datasets among clients, leading to more specialized models, higher accuracy, and lower communication costs. Other works \cite{ibnatta2022indoor, 9761235, guo2023fedpos} also integrated FL into the design of localization systems, demonstrating its potential for an efficient collaborative learning. 

In the context of VLC systems, FL has been successfully applied to VLC positioning systems to address privacy preservation and collaborative model training challenges. For instance, \cite{wei2023privacy} proposed a privacy-preserving cooperative VLC positioning scheme based on FL for nonstationary environments. Their approach employs a Cooperative Visible-light Positioning Network (CVPosNet) that enables global model training without sharing private user data, achieving improved positioning accuracy and generalization capability in dynamic lighting conditions. Hybrid VLC/RF architectures have also been adopted to offload model updates onto the visible spectrum, improving uplink bandwidth efficiency \cite{huang2021wireless}. 

Despite these advancements, VLC-FL systems require strict line-of-sight (LOS) to LEDs, which limits client availability and leads to uneven data contributions. Uplink communication via RF or Infrared (IR) links also suffers from bandwidth constraints, slowing global aggregation and hindering scalability in dynamic environments. Moreover, FL-trained models often become overly tailored to the specific data distributions of participating clients. This hinders generalization to unfamiliar devices and unseen deployment environments \cite{rev1p2}. These limitations expose critical performance disparities across heterogeneous deployments and underscore the need for architectural enhancements, such as TL, to improve cross-area robustness without extensive retraining or additional data collection.

\subsection{CL for Indoor Localization}
CL allows incremental model updates as new data arrive, suiting environments with dynamic signal characteristics \cite{wickramasinghe2023continual}. Recent surveys, such as \cite{van2022three} provide empirical insights to CL techniques. A key issue in CL is \textit{catastrophic forgetting}, where a model loses previously acquired knowledge when trained on new tasks. Overcoming this problem typically requires additional memory capacity, complex update strategies, and increased computational costs, which may compromise CL’s suitability for resource-limited environments. Various CL strategies have been proposed to address this issue. As outlined in \cite{de2021continual}, task-incremental learning strategies fall into three broad categories: replay-based (involving storage and reprocessing of prior samples), regularization-based (enforcing constraints on parameter shifts), and parameter isolation (dedicating separate parameter subsets for different tasks). 
Each approach brings trade-offs—replay increases memory and latency, regularization demands frequent updates, and parameter isolation reduces scalability. Similarly, \cite{masana2022class} investigated class-incremental learning for image classification, showing that techniques such as rehearsal and bias mitigation further complicate the computational and architectural design. The feasibility of CL depends not only on algorithmic design, but also on access to labeled data and computational resources, both commonly restricted in industrial environments. Furthermore, the need for frequent updates in response to new data can lead to variable energy consumption, complicating deployment of CL on energy-constrained devices such as battery-powered IoT nodes. These operational challenges limit the scalability of CL for industrial indoor localization, where systems must remain both adaptive and efficient \cite{wen2018efficient}. 

Within VLC systems, CL has been applied to address environmental dynamics such as lighting variation, mobility, and access point availability. Approaches include adaptive Kalman filtering \cite{eroglu2019adaptive} and RSS-based MIMO localization that adapt positioning in real-time \cite{datnovel}. Broad learning systems with self-attention have also shown promise for continuous model updates \cite{zhao2025self}. However, abrupt optical channel changes and the non-stationary nature of VLC signals intensify catastrophic forgetting. Moreover, real-time adaptation requires frequent computation and memory access, conflicting with VLC’s hardware constraints. These factors hinder CL’s effectiveness in VLC environments, where fast yet efficient adaptation is essential. In this context, TL offers a more lightweight and scalable alternative by enabling fast generalization to new environments using minimal data and limited fine-tuning.

\subsection{Alternative Adaptation Techniques for Indoor Localization}
Several recent studies have explored alternative adaptation techniques such as domain-adversarial training and meta-learning to address domain shifts across environments. Domain-adversarial approaches learn domain-invariant representations by introducing a discriminator that aligns feature distributions between source and target environments \cite{ganin2016domain, zhang2023domain}. For WiFi CSI fingerprinting, recent works employ graph convolutional networks with adversarial alignment to leverage large crowdsourced datasets across multiple buildings \cite{shi2025mdaaloc}, achieving competitive accuracy. However, these methods typically require complex graph construction, substantial labeled or unlabeled target data for stable training, and additional optimization stages that increase computational cost and implementation complexity. Such requirements can be challenging in industrial VLC deployments, where measurement campaigns must be conducted under real production constraints and additional adversarial training stages increase implementation complexity and computational overhead.

Meta-learning approaches such as Model-Agnostic Meta-Learning (MAML) learn model initializations that enable rapid adaptation to new tasks \cite{owfi2023meta}. In wireless localization, these methods have been applied across numerous sites or combined with FL to support dynamic WiFi positioning \cite{etiabi2024femloc, cha2025meta}. Their effectiveness depends on exposure to many diverse environments during meta-training and on repeated task-level optimization, which can be difficult in industrial VLC scenarios due to the need for specialized meta-training procedures and the practical effort required for repeated data labeling and system calibration.

Compared with adversarial or meta-learning pipelines, which often involve additional optimization stages and more complex training procedures, TL offer computational simplicity while achieving substantial cross-domain performance recovery.

\subsection{TL for Indoor Localization}
While FL and CL have advanced the applicability of adaptive indoor positioning, their reliance on extensive local data, continual training and limited generalizability continue to constrain practical deployment, especially in optical domains where device connectivity and synchronization are limited. In contrast, TL offers a compelling alternative by leveraging pretrained models to enable rapid adaptation in new environments using minimal data and limited retraining. These results in faster deployment, reduced resource usage, and improved robustness across deployment areas \cite{iman2023review, ali2023transfer}. As discussed in \cite{ying2018transfer, pan2009survey}, the effectiveness of TL depends on understanding when, what, and how to transfer knowledge. These decisions are particularly important in industrial environments, where the trade-off between data collection costs, energy budgets and adaptation delays must be minimized. Compared to FL and CL, TL avoids the need for continuous client participation or memory-intensive model updates, making it more suitable for resource-constrained VLC systems. 

TL has demonstrated effectiveness in RF-based localization systems. For example, in Wi-Fi fingerprinting, fine‑tuning convolutional neural network (CNN) models on just 55\% of target data recovered over 95\% of full-data accuracy \cite{pan2008transfer}. Similarly, \cite{suwannaphong2024transfer} applied TL to BLE room classification across homes, significantly improving F1 scores in unseen settings, and UWB systems maintained sub-30 cm accuracy while reducing fingerprinting effort by 93\% \cite{adamkiewicz2021improving}. 

Despite this progress, TL remains underexplored in VLC based localization, where distinct optical-domain challenges persist. VLC offers advantages such as high bandwidth, EMI immunity, and energy efficiency through infrastructure reuse, yet most prior work emphasizes deterministic modeling or hardware mitigation. For example, \cite{zhuang2019visible} employed adaptive least squares and Kalman filtering to mitigate noise under fixed conditions, while \cite{zhang2023recent} utilized singular value decomposition (SVD) for efficient 3D positioning, achieving high accuracy but lacking generalization. \cite{rekkas2023artificial} noted that learning-based VLC methods still struggle with robustness and transferability. Furthermore, recent TL efforts in RF-based systems rarely account for VLC-specific factors like ambient light variation, photodiode-induced noise, or LED flicker. For instance, \cite{yang2024indoor} applied domain-adversarial learning to reduce Wi-Fi localization errors, but did not address interference in the optical domain. Similarly, \cite{ahmed2024unified} improved LoRa/Wi-Fi positioning via latent space alignment, without considering degradation modes specific to VLC. 

Our previous study has shown that TL-based VLC positioning can improve accuracy by up to 47\%, while reducing energy consumption and computational time by over 30\% relative to conventional training approaches \cite{11174705}. That study applied TL within a single industrial area to improve robustness under varying lighting conditions, demonstrating that VLC localization can adapt to environmental changes with reduced data requirements. In contrast, the present work addresses a different challenge: cross-area adaptation. We develop a TL framework that enables models trained in one deployment area to generalize to new areas, and introduce a new metric to systematically identify the most suitable source model without exhaustive testing.

These gaps motivate our work: a cross-area-aware TL framework for VLC localization, underpinned by (i) \emph{data-efficient fine-tuning} that cuts data collection costs by 60\% while preserving over 97\% of full-data performance, and (ii) a \emph{TES} metric to rank source models without exhaustive trials. Our approach is scalable—meaning it can accommodate additional deployment areas with minimal extra data or retraining—and energy-efficient, offering a lightweight, rapid adaptation solution tailored to industrial VLC environments.

\section{System Model and Problem Formulation}
\label{sec:sysm}

This section provides a structured approach for modeling and evaluating our deep neural network (DNN) configuration across different deployment areas, focusing on the challenge of TL and model generalization. 
DNNs are employed as the core regression architecture due to their proven effectiveness in VLC-based indoor positioning systems, where they can capture complex non-linear relationships between received signal strength (RSS) measurements and spatial coordinates that traditional ML approaches struggle to model effectively. Unlike conventional ML techniques such as linear regression or support vector machines, DNNs can learn hierarchical feature representations from raw RSS data, making them particularly suitable for handling the multipath effects, ambient light variations, and LED flicker characteristics inherent in VLC environments.
We further introduce the TES, a unified metric to guide the selection of the optimal source model for knowledge transfer. 

\subsection{System Model}


\begin{figure*}[!t]
  \centering
  \resizebox{1.0\linewidth}{!}{\includegraphics{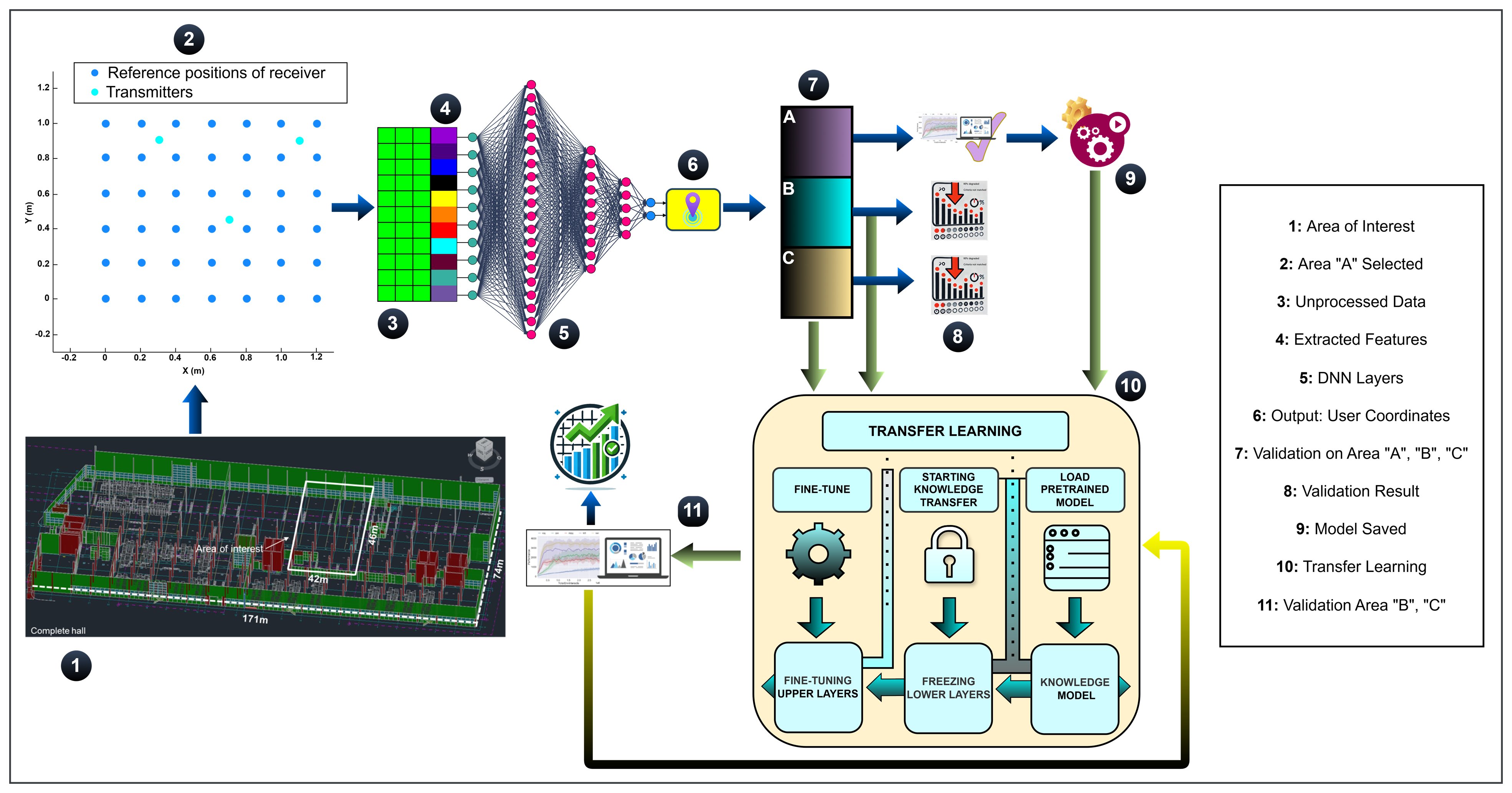}}
  \caption{System Model of the TL step.}
  \label{fig:sys}
\end{figure*}

Our goal is to develop a TL approach for predicting users’ coordinates (longitude and latitude) in a cross-area setting, using a VLC dataset collected from the factory hall of a Bosch plant in Blaichach, Germany. To support robust training and evaluation across varying deployment areas, the actual measurement area (area of interest shown in Fig.~\ref{fig:sys}) is partitioned into non-overlapping subareas (Areas~A, B, C, and D). We partitioned these areas based on their architectural diversity, including differences in layout dimensions, equipment configurations, and structural element distribution. These variations introduce distinct environmental conditions, creating a realistic and challenging testbed for evaluating cross-area model generalization (see Section~\ref{sec:setup} for more details). For each experimental round, one area (e.g., Area~D) is withheld as an unseen deployment area, while the remaining areas are used for model development and validation. This rotation process ensures comprehensive generalization assessment across all spatial contexts.

As illustrated in Fig.~\ref{fig:sys}, a DNN is trained on the dataset collected from one area (the “source” e.g., Area A) and initially validated on the validation dataset of the same area. To assess cross-area robustness, the trained model is then evaluated on datasets from a different area (the “target,” e.g., Area B or C). If the model’s validation performance on the target area (in terms of localization error and success rate) is significantly degraded compared to its performance on the source area, we adopt a TL approach to mitigate this cross-area generalization gap. Therefore, the model is fine-tuned on a small subset of the target area, freezing lower layers to preserve general feature representations. The target area is re-evaluated after fine-tuning to assess whether performance improved beyond the identified degradation. This process is repeated for all possible source-target pairs (A→B, A→C, B→A, B→C, C→A, C→B), providing a comprehensive assessment of cross-area generalization and adaptation for validation purposes within the known areas (A, B, C) before applying the selected model to the truly unseen area D.

\begin{figure*}[!t]
  \centering
  \resizebox{1.0\linewidth}{!}{\includegraphics{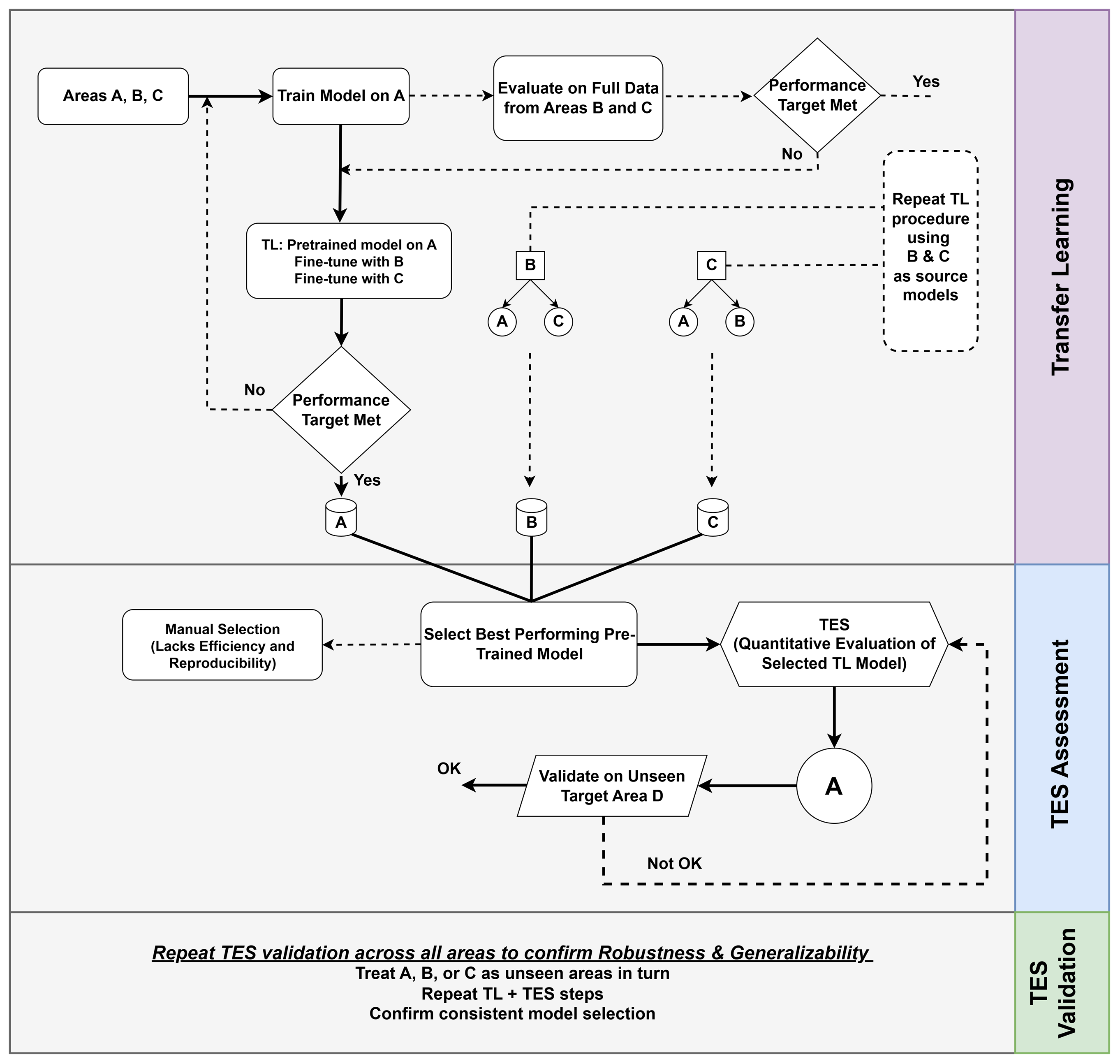}}
  \caption{Workflow of the proposed TL framework with TES assessment and validation.}
  \label{fig:TES}
\end{figure*}

To systematically identify the most effective transfer path, we compute a TES for each source model after fine-tuning on different target areas. Fig.~\ref{fig:TES} details this automated workflow, which replaces manual selection with a repeatable metric. The source area with the highest TES score is identified as the most reliable for knowledge transfer. The pre-trained model from this optimal source area is then validated on an unseen area (e.g., Area D) to assess its broader generalizability. By rotating the role of the unseen area across all subareas, we ensure that the selected source model maintains strong performance across diverse deployment areas when applied as a foundation for TL to new environments.


\subsection{Problem Formulation}

Let $\mathcal{A}=\{A,B,C\}$ be the set of source subareas and let $\mathcal{D}_i$ be the VLC dataset collected in area $i$;
\begin{equation}
    \mathcal{D}_i = \{(x^i_n, y^i_n)\}_{n=1}^{N_i}, 
\qquad i\in\mathcal{A},
\end{equation}

where \(x^i_n\in\mathbb{R}^d\) are the input features (VLC RSS readings) and \(y^i_n = (\ell^i_n,\phi^i_n)\in\mathbb{R}^2\) are the corresponding true coordinates, with \(\ell^i_n\) denoting the longitude and \(\phi^i_n\) the latitude.

Our objective is to build a DNN regressor $f_{\theta}:\mathbb{R}^d\rightarrow\mathbb{R}^2$ that, when trained on any source deployment area and fine-tuned on a small labeled subset of a target deployment area, achieves low localization error and high transferability across all subareas. To systematically guide the selection of the source model, we further integrate the \textit{TES}, which consolidates the accuracy and success rate of the localization into a unified metric. Thus, the problem formulation jointly considers (i) the effectiveness of TL and (ii) the identification of the most reliable source area through TES.

\subsubsection{Source Training \& Direct Evaluation}

For each source area $s\in\mathcal{A}$, we train an instance of the DNN with identical architecture and hyperparameters, but with weights initialized and optimized independently on the dataset of that area. This ensures comparability across source areas while avoiding any architectural bias. The parameters $\theta_s$ are obtained by minimizing a regression loss $\mathcal{L}$ (here mean squared error, MSE) on the full source dataset as:

\begin{equation}
  \theta_s 
  = \arg\min_{\theta}
    \;\mathcal{L}\bigl(f_{\theta}(X_s),\,Y_s\bigr),
  \label{eq:source_train}
\end{equation}

where
\[
  X_s = \{x^s_n\}_{n=1}^{N_s},
  \quad
  Y_s = \{y^s_n\}_{n=1}^{N_s}
\]

Here, $\theta_s$ denotes the learned network parameters (weights and biases) after training on area $s$. $N_s$ is the total number of training samples collected in area $s$. 
$f_{\theta}:\mathbb{R}^d\to\mathbb{R}^2$ is the DNN regressor, mapping each $d$‑dimensional RSS vector $x^s_n$ to a 2‑vector of predicted coordinates. We denote this prediction by;
\begin{equation}
    \hat y^s_n = f_{\theta}(x^s_n)\in\mathbb{R}^2,
\end{equation}

$\mathcal{L}\bigl(f_{\theta}(X_s),Y_s\bigr)$ is the MSE over the dataset $(X_s,Y_s)$, i.e.
\begin{equation}
    \mathcal{L}\bigl(f_{\theta}(X_s),Y_s\bigr)
      = \frac{1}{N_s}\sum_{n=1}^{N_s}\bigl\|f_{\theta}(x^s_n) - y^s_n\bigr\|_2^2,
\end{equation}

We then evaluate this pretrained model directly 
on each target area $t \in \mathcal{A}\setminus\{s\}$:
\begin{equation}
  E_{s\to t}
  = \mathcal{L}\bigl(f_{\theta_s}(X_t),\,Y_t\bigr),
  \label{eq:direct_eval}
\end{equation}
where
\[
  X_t = \{x^t_n\}_{n=1}^{N_t},
  \quad
  Y_t = \{y^t_n\}_{n=1}^{N_t},
\]

\subsubsection{Fine‑Tuning}

Direct evaluation (Eq.~\ref{eq:direct_eval}) already provides a measure of how well the pretrained model $f_{\theta_s}$ generalizes to a target area $t \ne s$. If this evaluation reveals a substantial degradation in localization performance compared to validation on the source area, we then proceed with fine-tuning to adapt the model. To do so, we fine‑tune $f_{\theta_s}$ using a small randomly sampled labeled subset of the target dataset:
\begin{equation}
    \mathcal{D}_t^{(\mathrm{FT})}
= \bigl\{(x^t_n,y^t_n)\bigr\}_{n=1}^{M}, 
\quad M = \rho\,N_t,\quad \rho\in[0,1],
\end{equation}
where $M$ is the number of fine-tuning samples, $\rho$ is the sampling ratio, and $N_t = |\mathcal{D}_t|$. We experimented with different values of $\rho$ (10\%, 20\%, 40\%, 60\%, 100\%) to study the tradeoff between performance and data availability. Following the evaluation in Section~\ref{ft}, we select $\rho=40\%$ as the fine-tuning ratio used in all subsequent experiments. The subset $\mathcal{D}_t^{(\mathrm{FT})}$ is drawn at random from $\mathcal{D}_t$.

To preserve the general representations learned from the source area, only the final $k$ layers of the DNN are unfrozen and updated using $\mathcal{D}_t^{(\mathrm{FT})}$, while the lower layers remain fixed. The motivation and impact of this fine-tuning strategy under different target-data regimes are further analyzed in Section~\ref{ftstrategy}. The fine-tuning operation is denoted by $\psi(\theta_s, \mathcal{D}_t^{(\mathrm{FT})})$, representing gradient updates to the unfrozen parameters. The adapted model parameters are then:
\begin{equation}
    \theta_{s\to t}
= \psi\bigl(\theta_s,\mathcal{D}_t^{(\mathrm{FT})}\bigr),
\end{equation}

Following fine-tuning, we evaluate the adapted model on the full target dataset to measure its localization error and success rate:
\begin{align}
E_{s\to t}^{(\mathrm{FT})}
&=\mathcal{L}\bigl(f_{\theta_{s\to t}}(X_t),\,Y_t\bigr), \\
S_{s\to t}^{(\mathrm{FT})}
&= \frac1{N_t}\sum_{n=1}^{N_t}
\mathbf{1}\bigl(\|\hat y_n - y^t_n\|\le \delta\bigr),
\label{eq:SR}
\end{align}
where $\delta$ is a predefined success threshold and $\hat{y}_n = f_{\theta_{s\to t}}(x^t_n)$ is the model prediction.

\subsubsection{Transfer Directions}

Now let's define the set of all ordered source–target pairs $\mathcal{T}$:
\begin{equation}
    \mathcal{T}
=\{(s,t)\mid s,t\in\mathcal{A},\,s\neq t\},
\end{equation}

so there are six transfer directions: $A\!\to\!B$, $A\!\to\!C$, $B\!\to\!A$, etc.  For each $(s,t)\in\mathcal{T}$ we obtain a fine‑tuned model $\theta_{s\to t}$ and corresponding error $E_{s\to t}^{\mathrm{(FT)}}$. 

\subsubsection{Transfer Evaluation Score (TES)}
\label{subsubsec:TES}

To systematically evaluate and rank these \(6\) candidates, we define the \textit{TES}, a composite metric that reflects both post-transfer localization accuracy and robustness. For each transfer scenario, we compute the minimum and maximum localization errors \(E_{\text{min}}\), \(E_{\text{max}}\) and minimum and maximum success rates \(S_{\text{min}}\), \(S_{\text{max}}\) after fine-tuning as expressed below: Importantly, $E_{\min}, E_{\max}, S_{\min}, S_{\max}$ are computed independently for each transfer scenario $(s,t)$, so that the normalization in Eq.~\ref{eq:norm} reflects the performance range within that specific scenario, without being influenced by other transfers.

\begin{equation}
\label{eq:minmax}
\left\{
  \begin{aligned}
    &E_{\min} = \min_{(s,t)\in\mathcal{T}} E_{s \to t}^{(\mathrm{FT})} \\
    &E_{\max} = \max_{(s,t)\in\mathcal{T}} E_{s \to t}^{(\mathrm{FT})} \\
    &S_{\min} = \min_{(s,t)\in\mathcal{T}} S_{s \to t}^{(\mathrm{FT})} \\
    &S_{\max} = \max_{(s,t)\in\mathcal{T}} S_{s \to t}^{(\mathrm{FT})}
  \end{aligned}
\right.,
\end{equation}

\noindent We then compute the normalized error and success‑rate. Normalization is applied so that both error and success-rate values are scaled to the same [0,1] range, regardless of their absolute magnitudes. This allows TES to combine them fairly in a single composite score, avoiding bias toward metrics with larger numerical ranges. 

\begin{align}
\label{eq:norm}
  \mathrm{NE}_{s\to t}
  = \frac{E_{s\to t}^{(\mathrm{FT})} - E_{\min}}{E_{\max} - E_{\min}}, \\
  \mathrm{NSR}_{s\to t}
  = \frac{S_{s\to t}^{(\mathrm{FT})} - S_{\min}}{S_{\max} - S_{\min}},
\end{align}

For each \((s,t)\), the per‑target TES is defined as:
\begin{equation}
\label{eq:tes}
\mathrm{TES}_{s\to t}
= \alpha\,(1-\mathrm{NE}_{s\to t}) \;+\; (1-\alpha)\,(\mathrm{NSR}_{s\to t}),
\end{equation}
where \(\alpha \in [0, 1]\) controls the trade-off between localization error and success rate. In our experiments we set $\alpha = 0.5$, giving equal weight to both metrics. This choice reflects the fact that both metrics are equally important for evaluating transferability: error captures precision, while success rate captures robustness. To verify the robustness of this choice, we performed a sensitivity analysis of TES for $\alpha \in [0.3,\,0.7]$, as reported in Section~\ref{subsec:tes-validation}.

After computing $\mathrm{TES}_{s\to t}$ for every source–target pair $(s,t)\in\mathcal{T}$, we group the scores by their originating source area. Each source area contributes two transfer scenarios (e.g., A→B and A→C), and the TES values from these scenarios are averaged to produce one representative score. This consolidated score reflects how consistently a source transfers knowledge across all its targets, revealing which source area is the strongest candidate for pretraining and deployment.


We then select the source area with the highest averaged score:
\begin{equation}
    s^* = \underset{s \in \mathcal{A}}{\arg\max}\;\mathrm{TES}_{s},
\end{equation}

Here, $s^*$ identifies the source area whose model demonstrates the most reliable and balanced transfer performance across all possible target areas, making it the best candidate for deployment and for subsequent evaluation on unseen areas.

\subsubsection{Validation on Unseen Areas}

To evaluate the generalization capability of the selected model, we take the source model identified via TES from the best-performing source area $s^*$ and test it on a held-out subarea $D$, which was excluded from training and model selection but retained for evaluation. The localization error and success rate on area D are computed as:

\begin{align}
E_{s^* \to D}
&= \mathcal{L}\bigl(f_{\theta_{s^*}}(X_D),\,Y_D\bigr),
\\
S_{s^* \to D}
&= \frac{1}{N_D}\sum_{n=1}^{N_D}
\mathbf{1}\bigl(\|\hat{y}_n - y^D_n\| \le \delta\bigr),
\end{align}
where \( \hat{y}_n = f_{\theta_{s^*}}(x^D_n) \) is the predicted coordinate, \( y^D_n \) is the ground-truth location, and \( \delta \) is a pre-defined success threshold. The indicator function \( \mathbf{1}(\cdot) \) returns 1 if the prediction is within \( \delta \) of the ground truth, and 0 otherwise.

This evaluation quantifies the ability of the selected model to generalize to unseen environments. To further validate the reliability and consistency of the TES-based selection strategy, we rotate the role of the unseen area among all available subareas \( \mathcal{A} = \{A, B, C, D\} \). In each iteration, one area (e.g., A) is excluded from training and used solely for validation. The TES is computed using the remaining areas (e.g., B, C, and D). This simulates a realistic deployment scenario where the model must generalize to a new environment using transfer experience from previously explored regions, and serves as a validation step to confirm the robustness and consistency of the TES-based model selection.

\section{Algorithm development and implementation}
\label{sec:algorithm}

This section presents a two-stage algorithmic framework designed to enable efficient and scalable model transfer for VLC-based indoor localization. The primary objective is to promote cross-area generalization by systematically evaluating when and how TL should be applied across spatial subareas.

Our approach consists of two interconnected components: (i) a cross-area model training and evaluation procedure that establishes baseline performance across all deployment areas, and (ii) a TES-based selection and validation mechanism that identifies optimal transfer configurations. These components work synergistically to ensure robust localization performance while minimizing computational overhead.

\begin{algorithm}
\fontsize{8pt}{9pt}\selectfont
\caption{Cross-Area Model Training and Transfer Capability Evaluation}
\label{alg:training_evaluation}
\SetAlgoLined
\SetKwInOut{Input}{Input}
\SetKwInOut{Output}{Output}

\Input{$\mathcal{A} = \{A, B, C\}$; /* Set of available areas */}
\Input{$\{\mathcal{D}_i\}_{i \in \mathcal{A}}$; /* Datasets for each area */}
\Input{$\rho$; /* Fine-tuning sampling ratio */}
\Input{$k$; /* Number of layers to unfreeze */}
\Input{$\delta$; /* Success threshold */}
\Output{
$\{\theta_{s \to t}\}_{(s,t) \in \mathcal{T}}$: /* All fine-tuned models */ \\
$\{E_{s \to t}^{(\mathrm{FT})}\}_{(s,t) \in \mathcal{T}}$: /* Transfer errors */ \\
$\{S_{s \to t}^{(\mathrm{FT})}\}_{(s,t) \in \mathcal{T}}$: /* Transfer success rates */}

\BlankLine
\textbf{Train Models on All Source Areas} \\
\For{each source area $s \in \mathcal{A}$}{
    Train DNN model $f_{\theta_s}$ on $\mathcal{D}_s$: \\
    \quad $\theta_s \gets \arg\min_{\theta} \mathcal{L}(f_\theta(X_s), Y_s)$
}

\BlankLine
\textbf{Evaluate All Source-Target Transfer Pairs} \\
Initialize $\mathcal{T} = \{(s,t) \mid s,t \in \mathcal{A}, s \neq t\}$ \\
\For{each $(s,t) \in \mathcal{T}$}{
    \textbf{Direct Cross-Area Evaluation:} \\
    \quad $E_{s \to t} = \mathcal{L}(f_{\theta_s}(X_t), Y_t)$ \\
    \quad $S_{s \to t} = \frac{1}{N_t} \sum_{n=1}^{N_t} \mathbf{1}(\|\hat{y}_n^t - y_n^t\| \le \delta)$ \\
    
    \textbf{Source Area Baseline:} \\
    /* Used only for comparison to decide if TL is needed */\\
    \quad $E_{s \to s} = \mathcal{L}(f_{\theta_s}(X_s), Y_s)$ \\
    \quad $S_{s \to s} = \frac{1}{N_s} \sum_{n=1}^{N_s} \mathbf{1}(\|\hat{y}_n^s - y_n^s\| \le \delta)$ \\

    \textbf{Apply TL if Performance Degrades:} \\
    \If{$E_{s \to t} > E_{s \to s}$ \textbf{or} $S_{s \to t} < S_{s \to s}$}{
        Sample fine-tuning subset: $\mathcal{D}_t^{(\mathrm{FT})} \subset \mathcal{D}_t$ with $|\mathcal{D}_t^{(\mathrm{FT})}| = \rho \cdot N_t$ \\
        Initialize $\theta \gets \theta_s$ \\
        Freeze lower $(L-k)$ layers of $f_\theta$ \\
        Fine-tune: $\theta_{s \to t} \gets \psi(\theta_s, \mathcal{D}_t^{(\mathrm{FT})})$ \\
        
        Re-evaluate with fine-tuned model: \\
        \quad $E_{s \to t}^{(\mathrm{FT})} = \mathcal{L}(f_{\theta_{s \to t}}(X_t), Y_t)$ \\
        \quad $S_{s \to t}^{(\mathrm{FT})} = \frac{1}{N_t} \sum_{n=1}^{N_t} \mathbf{1}(\|\hat{y}_n - y_n^t\| \le \delta)$
    }
    \Else{
        $E_{s \to t}^{(\mathrm{FT})} = E_{s \to t}$, $S_{s \to t}^{(\mathrm{FT})} = S_{s \to t}$ \\
        $\theta_{s \to t} = \theta_s$
    }
}

\BlankLine
\Return $\{\theta_{s \to t}\}_{(s,t) \in \mathcal{T}}$, $\{E_{s \to t}^{(\mathrm{FT})}\}_{(s,t) \in \mathcal{T}}$, $\{S_{s \to t}^{(\mathrm{FT})}\}_{(s,t) \in \mathcal{T}}$
\end{algorithm}

We begin by defining a set of spatial subareas $\mathcal{A} = \{A, B, C\}$, each associated with its own dataset $\mathcal{D}_i$. For every source area $s \in \mathcal{A}$, a DNN model $f_{\theta_s}$ is trained by minimizing a supervised loss function $\mathcal{L}$ over the corresponding dataset $\mathcal{D}_s$. The resulting model parameters $\theta_s$ serve as source models for potential transfer to other target areas, forming the basis of our TL strategy.

Building upon this foundation, Algorithm~\ref{alg:training_evaluation} evaluates all possible source–target area  combinations $(s, t)$, where $s \ne t$. For each pair, (i) a direct cross-area inference is performed using the source model $f_{\theta_s}$ on the target dataset $\mathcal{D}_t$, yielding an initial localization error $E_{s \to t}$ and success rate $S_{s \to t}$. (ii) A baseline reference is computed by evaluating $f_{\theta_s}$ on its own training data $\mathcal{D}_s$, producing $E_{s \to s}$ and $S_{s \to s}$. These evaluations establish a performance benchmark for each source model on its home area, which is then used to judge how well it generalizes to new environments. Whenever the performance on a target area falls short of this benchmark—that is, if the localization error on the target dataset exceeds the source baseline (\( E_{s \to t} > E_{s \to s} \)) or the success rate drops below it (\( S_{s \to t} < S_{s \to s} \))—the algorithm concludes that simple cross‑area inference is insufficient. At this point, the algorithm triggers TL to help the model adapt to the new area while retaining the useful representations learned from the source.

In the fine‑tuning stage, a small subset of the target dataset, $\mathcal{D}_t^{(\mathrm{FT})}$, is randomly sampled, containing a fraction $\rho$ of the total data. This subset is used to update only the top $k$ layers of the network, while the lower $(L-k)$ layers remain frozen to preserve general feature representations. The resulting adapted model $\theta_{s \to t}$ is then re‑evaluated to produce updated metrics: localization error $E_{s \to t}^{(\mathrm{FT})}$ and success rate $S_{s \to t}^{(\mathrm{FT})}$. By repeating this process for every source–target pair, the algorithm generates a comprehensive set of fine‑tuned models and performance metrics. These outputs form the input to Algorithm~\ref{alg:tes_selection_validation} which implements a selection procedure that identifies the optimal source–target area configuration using the TES. This metric jointly accounts for localization error and success rate, providing a balanced criterion for identifying the most effective TL setup.

\begin{algorithm}
\fontsize{8pt}{9pt}\selectfont
\caption{TES-Based Source Model Selection and Unseen Area Validation}
\label{alg:tes_selection_validation}
\SetAlgoLined
\SetKwInOut{Input}{Input}
\SetKwInOut{Output}{Output}

\Input{$\mathcal{T} = \{(s,t) \mid s,t \in \mathcal{A}, s \neq t\}$; /* All source-target pairs */}
\Input{$\{E_{s \to t}^{(\mathrm{FT})}\}_{(s,t) \in \mathcal{T}}$; /* Fine-tuned localization errors */}
\Input{$\{S_{s \to t}^{(\mathrm{FT})}\}_{(s,t) \in \mathcal{T}}$; /* Fine-tuned success rates */}
\Input{$\alpha \in [0,1]$; /* Error-success trade-off parameter */}
\Input{$\{\theta_s\}_{s \in \mathcal{A}}$; /* Source models trained on A,B,C */}
\Input{$\mathcal{D}_D$; /* Unseen area dataset */}
\Input{$\delta$; /* Success threshold */}

\Output{
$s^*$: /* Optimal source area */ \\
$\theta_{s^*}$: /* Best source model */ \\
$E_{s^* \to D}, S_{s^* \to D}$: /* Performance on unseen area */ }

\BlankLine
\textbf{Compute Global Min–Max Statistics} \\
$E_{\min} \gets \min_{(s,t)} E_{s \to t}^{(\mathrm{FT})}$, 
$E_{\max} \gets \max_{(s,t)} E_{s \to t}^{(\mathrm{FT})}$ \\
$S_{\min} \gets \min_{(s,t)} S_{s \to t}^{(\mathrm{FT})}$, 
$S_{\max} \gets \max_{(s,t)} S_{s \to t}^{(\mathrm{FT})}$

\BlankLine
\textbf{Compute TES for Each Pair} \\
\For{each $(s,t) \in \mathcal{T}$}{
    $\mathrm{NE}_{s \to t} \gets \frac{E_{s \to t}^{(\mathrm{FT})} - E_{\min}}{E_{\max} - E_{\min}}$ \\
    $\mathrm{NSR}_{s \to t} \gets \frac{S_{s \to t}^{(\mathrm{FT})} - S_{\min}}{S_{\max} - S_{\min}}$ \\
    $\mathrm{TES}_{s \to t} \gets \alpha (1 - \mathrm{NE}_{s \to t}) + (1-\alpha)(\mathrm{NSR}_{s \to t})$
}

\BlankLine
\textbf{Aggregate TES by Source Area} \\
\For{each $s \in \mathcal{A}$}{
    $\mathrm{TES}_s \gets \frac{1}{2} \sum_{t \neq s} \mathrm{TES}_{s \to t}$
}

\BlankLine
\textbf{Select Best Source Area} \\
$s^* \gets \arg\max_{s \in \mathcal{A}} \mathrm{TES}_s$ \\
$\theta_{s^*} \gets \theta_{s^*}$ /* Retain pretrained model from $s^*$ */

\BlankLine
\textbf{Validate on Unseen Area} \\
Extract $X_D, Y_D$ from $\mathcal{D}_D$ \\
$E_{s^* \to D} \gets \mathcal{L}(f_{\theta_{s^*}}(X_D), Y_D)$ \\
$S_{s^* \to D} \gets \frac{1}{N_D} \sum_{n=1}^{N_D} \mathbf{1}(\| \hat{y}_n - y_n^D \| \le \delta)$ \\
\quad where $\hat{y}_n = f_{\theta_{s^*}}(x_n^D)$

\BlankLine
\Return $s^*$, $\theta_{s^*}$, $E_{s^* \to D}$, $S_{s^* \to D}$
\end{algorithm}

Algorithm~\ref{alg:tes_selection_validation} begins by considering all valid source–target area pairs $(s, t) \in \mathcal{T}$, where $s \ne t$, and their corresponding fine-tuned models $\theta_{s \to t}$. Using the localization error $E_{s \to t}^{(\mathrm{FT})}$ and success rate $S_{s \to t}^{(\mathrm{FT})}$ which are computed during prior evaluations, the algorithm normalizes these values using min–max normalization to ensure comparability across different source–target scenarios.
For each evaluated pair, the normalized error $\mathrm{NE}_{s \to t}$ and normalized success rate $\mathrm{NSR}_{s \to t}$ are computed, and a pairwise TES score $\mathrm{TES}_{s \to t}$ is calculated using the weighted formulation in Eq.~\ref{eq:tes}. Once all TES values are computed, the TES scores are aggregated by source area: for each source $s$, the two TES values corresponding to its target areas (e.g., $A \to B$ and $A \to C$) are averaged to produce an overall score $\mathrm{TES}_{s}$. The source area with the highest aggregated TES, denoted by $s^*$, is then selected as the optimal source for deployment. Finally, this chosen source model $\theta_{s^*}$ is validated on the unseen area dataset $\mathcal{D}_D$ to assess its true generalization capability. The evaluation uses the same metrics (localization error and success rate) to ensure consistency.

By aggregating TES scores per source and validating only the top-ranked model on unseen data, this updated framework avoids overfitting to a single transfer direction and ensures that the selected model represents a broadly transferable solution for VLC-based indoor localization.


\section{Experimental Setup and Data Collection} 
\label{sec:setup}

The proposed VLC-based localization framework is based on measured data collected in a realistic industrial environment. This section outlines the VLC transceiver design, the measurement infrastructure deployed within the BOSCH facility, and the process used to gather signal strength readings. These measurements provide a representative basis for model training and evaluation under practical deployment conditions.

\subsection{VLC Transceiver System}
\subsubsection{Architecture}

\begin{figure}[!t]
    \centering
    \includegraphics[scale=0.36]{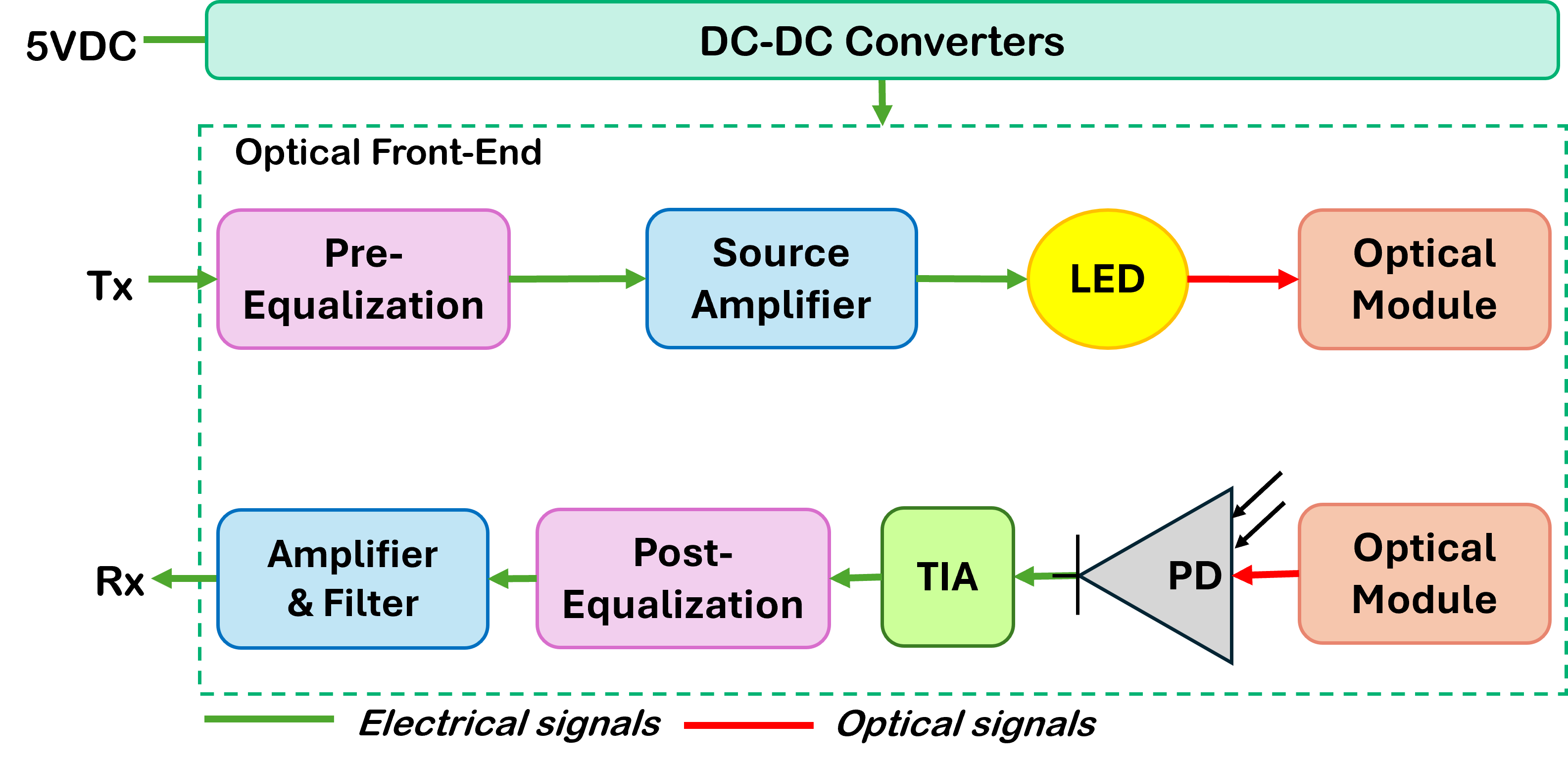}
    \caption{Architecture of the implemented VLC transceiver.}
    \label{fig:Architecture of the developed VLC transceiver}
\end{figure}

The transceiver is composed of three main subsystems: the power supply, the transmitter (Tx), and the receiver (Rx), each designed to ensure high signal fidelity and robustness across varying channel conditions. As illustrated in Fig.~\ref{fig:Architecture of the developed VLC transceiver}, it features an optical front-end powered by a regulated 5V DC input, stabilized via dedicated DC-DC converters. These converters generate multiple supply rails required at various stages of the transceiver, including the biasing of the photodetector (PD). 

On the Tx side, electrical signals are first processed through a pre-equalization stage to compensate for channel-induced distortions and bandwidth limitations of the LEDs. Since LEDs can only respond effectively to signals within a limited frequency range, signals containing higher frequency components may be distorted or attenuated. The pre-equalization stage mitigates these effects, enhancing signal fidelity before optical transmission. These equalized signals are then amplified by a source amplifier to meet the current requirements of the LED, which converts the electrical signal into an optical signal. 

After free space propagation, the modulated optical signal is collected at the Rx side. The incoming optical signals are captured by a PD and converted into electrical currents. These currents are converted into voltage signals using a transimpedance amplifier (TIA), followed by a post-equalization block to compensate for distortions introduced by the Rx chain. The signal quality is further refined by an amplifier and filtering stage before being forwarded for downstream processing. The use of both pre- and post-equalization stages enhances end-to-end system performance by mitigating distortions across the optical wireless channel. The signal flow is shown in the diagram, where black arrows denote electrical signals and red arrows represent optical signals.

\subsubsection{Specifications} 

The transceiver unit is powered via a USB-C female connector, providing a convenient and standardized interface. The unit requires a 12~V supply at 1.2~A, resulting in a total power consumption of approximately 14.4~W. This ensures stable operation of both the optical front-end components and the digital control circuitry. Internally, dedicated DC-DC converters regulate the input voltage down to 5~V levels required by the LED drivers, photodiode amplifiers, and microcontrollers. The use of USB-C facilitates integration with commercially available power sources and provides robust mechanical durability for repeated connections in experimental settings.

The Tx employs a LED driven by a current signal with an amplitude of 200~mA and an operational variation range of ± 100~mA. This range allows modulation of the light intensity while minimizing nonlinear distortion and preventing thermal damage to the LED. The emitted optical power is centered around 500~mW with a tolerance of ± 250~mW, balancing the need for strong signal propagation with eye safety limits and thermal management considerations. The LED has a full angular emission width of 60° at half maximum power, allowing a moderately wide illumination cone to cover the target area while limiting spillover into adjacent zones. The selected peak wavelength of 450~nm, with an operational range of 450-650~nm and a spectral width of 40~nm, aligns with the high sensitivity region of standard photodiodes and ensures compatibility with commercially available optical filters. The Tx output is connected via a SubMiniature version A (SMA) interface, providing a secure and low-loss connection to ensure robust coupling.

On the Rx side, four PDs have been put in parallel to increase the sensitive area of the Rx. The signals produced by these PDs are then summed up to get a single signal to process and send to the VNA. These PDs already have an integrated optical lens which limits the full angle at half-sensitivity to 60° but allows to collect more optical power so that they have an effective sensitive area of 150 mm² whereas their actual sensitive area is 26.4 mm². Eventually, our VLC transceiver thus have a total sensitive area of 600 mm², with a photosensitivity at 400 nm. The spectral sensitivity of the Rx spans from 380nm to 780nm, effectively covering the visible light spectrum. At the target wavelength of 450nm, the PDs provide a photosensitivity of 0.3A/W, ensuring sufficient conversion of the incident light into electrical current, supporting accurate signal detection even under low-light conditions. 

Together, these specifications summarized in Table~\ref{tab:vlc_specs} enable a robust and tunable VLC system capable of capturing high-fidelity signal strength measurements across varying line-of-sight conditions, making it suitable for accurate indoor localization tasks in complex industrial environments.

\begin{table}[!t]
\centering
\caption{VLC Transceiver Specifications.}
\label{tab:vlc_specs}
\resizebox{8cm}{!}{   
\begin{tabular}{ll}
\toprule
\multicolumn{2}{c}{\textbf{Power Supply}} \\
\midrule
Supply voltage         & 12 V \\
Supply current         & 1.2 A \\
Power supply interface & USB-C female \\
\addlinespace[5pt]
\multicolumn{2}{c}{\textbf{Tx}} \\
\midrule
Driving current signal range & 200 mA $\pm$ 100 mA \\
Transmitted optical power range & 500 mW $\pm$ 250 mW \\
Full angle at half power & 60$^\circ$ \\
Wavelength of operation & 450 nm (450--650 nm) \\
Spectral width & 40 nm \\
Tx chain interface & SMA \\
\addlinespace[5pt]
\multicolumn{2}{c}{\textbf{Rx}} \\
\midrule
Effective sensitive area & 600 mm$^2$ \\
Photosensitivity width & 400 nm (380--780 nm) \\
Photosensitivity at 450 nm & 0.3 A/W \\
Full angle at half sensitivity & 60$^\circ$ \\
Rx chain interface & SMA \\
\bottomrule
\end{tabular}}
\end{table}

\subsection{VLC Measurement System}
\label{sec:vlc_meas_system}

\begin{figure}[!t]
    \centering
    \includegraphics[scale=0.6]{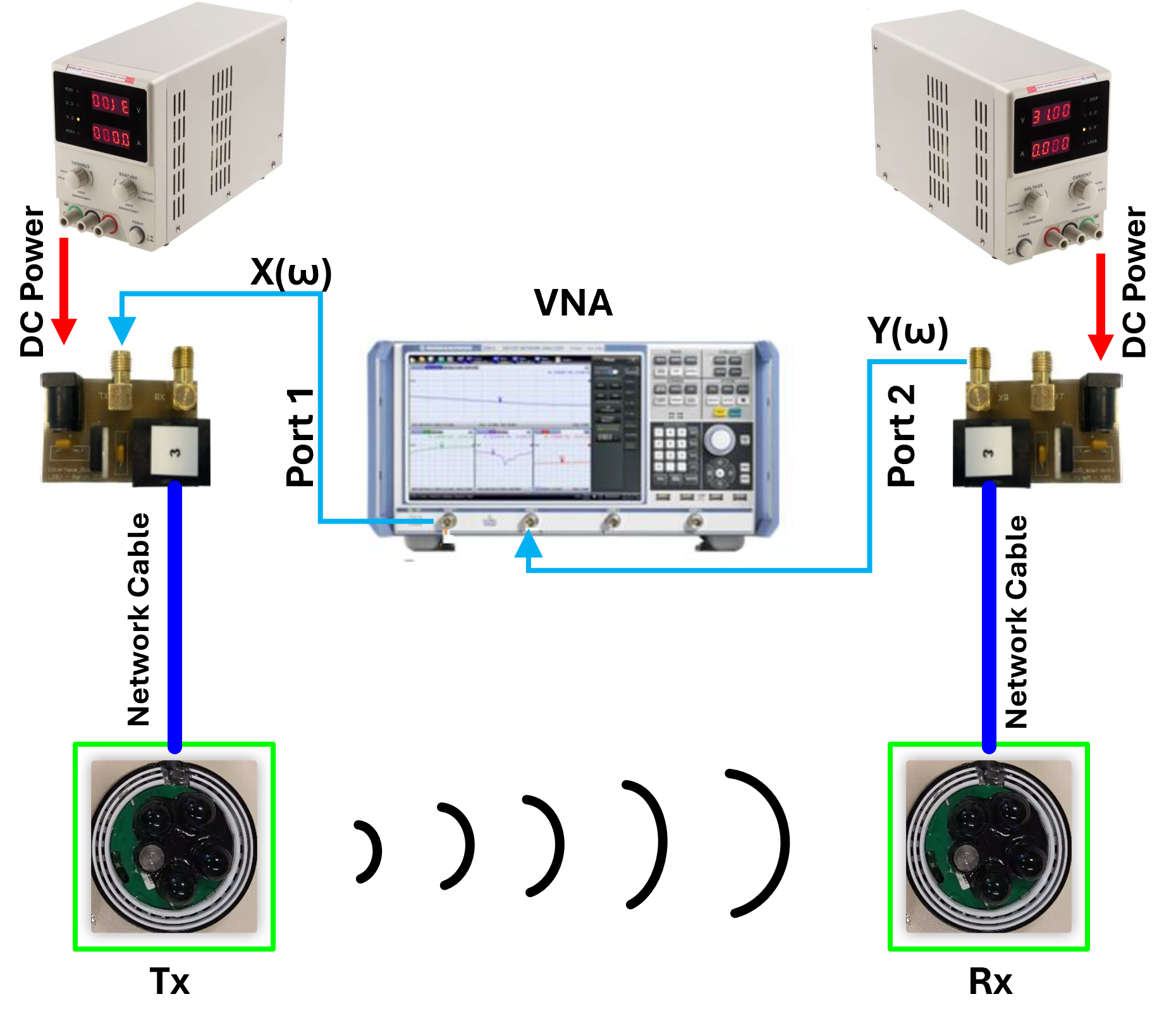}
    \caption{VLC measurement system.}
    \label{fig:vlc measurement setup}
\end{figure}

The complete setup, illustrated in Fig.~\ref{fig:vlc measurement setup}, was deployed at the BOSCH manufacturing facility. It consists of two VLC transceiver units (Tx and Rx), a dual-port vector network analyzer (VNA), two signal and power converter boards, and two regulated DC power supplies.

The VNA serves as both the signal generator and measurement instrument. During measurements, it generates a frequency-swept sinusoidal signal at Port~1, which modulates the optical carrier transmitted through the VLC system. This signal is first routed through the Tx-side converter board, where it is combined with a 12V DC supply from the external power module. The merged signal travels via a network cable to the Tx module, where the electrical signal drives the LED through the pre-equalization and driver stages described above.

The modulated optical signal propagates through the industrial environment and is captured by the Rx module. The received light is converted into an electrical signal using a photodiode-based front-end, and further amplified through the post-equalization chain. The resulting signal is returned via the second converter board, which separates the RF and DC components before forwarding the recovered signal to Port~2 of the VNA.

The use of converter boards enables a unified interface for signal and power transmission, simplifying cabling and improving measurement stability. The VNA records both the transmitted and received signals to compute signal strength measurements across different locations. These measurements, representing frequency-dependent power attenuation, serve as location-specific fingerprints and form the basis of the dataset used for training and evaluating the proposed TL based localization framework.

The system architecture allows for precise synchronization and repeatable operation of the Tx and Rx chains, ensuring that the collected signal strength measurements are consistent and representative of the actual indoor propagation dynamics. Such consistency is essential for constructing reliable datasets used in model adaptation and TL scenarios in complex industrial environments.

\subsection{Data Collection} 

\begin{figure*}[!t]
    \centering
    \includegraphics[scale=0.8]{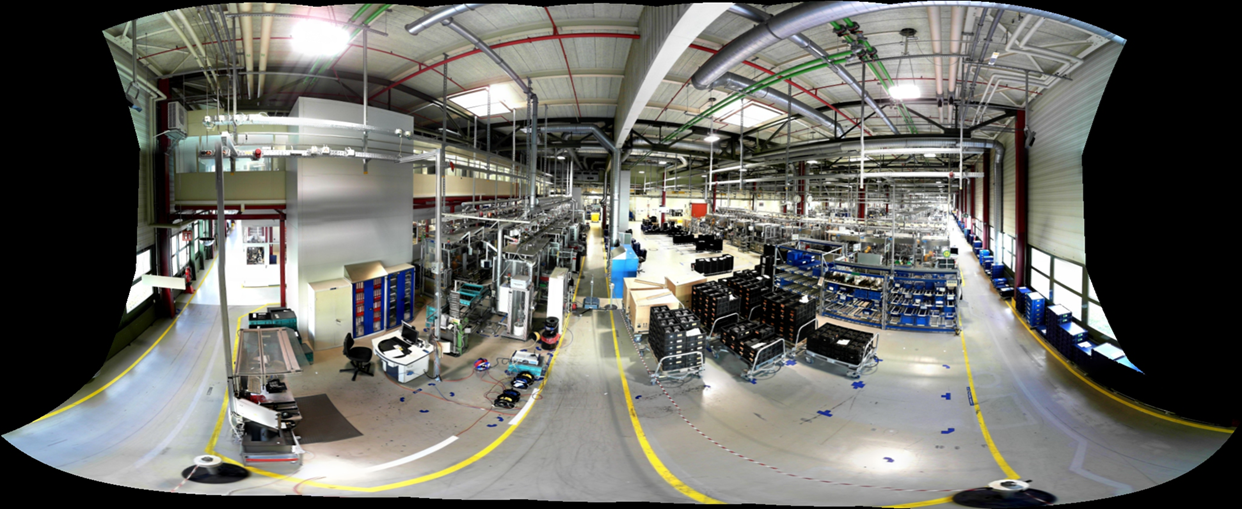}
    \caption{Composite picture from the Tx towards the measurement area in the industry hall at BOSCH in Blaichach.}
    \label{fig:compo}
\end{figure*}

The data collection took place in the industrial hall at the BOSCH factory in Blaichach, Germany. 
The complete hall has a dimension of 74 m $\times$ 171 m $\times$ 11 m, however, the measurement campaign focused on a representative measurement area, measuring 46 m $\times$ 42 m $\times$ 11 m. This segment contains one production line and a large storage area, as shown in the picture from the Tx shown in Fig. \ref{fig:compo}. 
The structural composition of the hall includes materials such as steel, concrete, glass, and aluminum. The external boundaries are formed by a combination of reinforced concrete walls, metal cladding, and glass surfaces. Inside, concrete columns are uniformly distributed along the interior. The ceiling-level part of the hall is densely equipped with metallic infrastructure such as ventilation ducts, suspended girders, and cable trays installed near the ceiling.


\begin{figure*}[!t]
    \centering
    \subfloat[]{%
        \raisebox{0.4cm}{\includegraphics[scale=0.4]{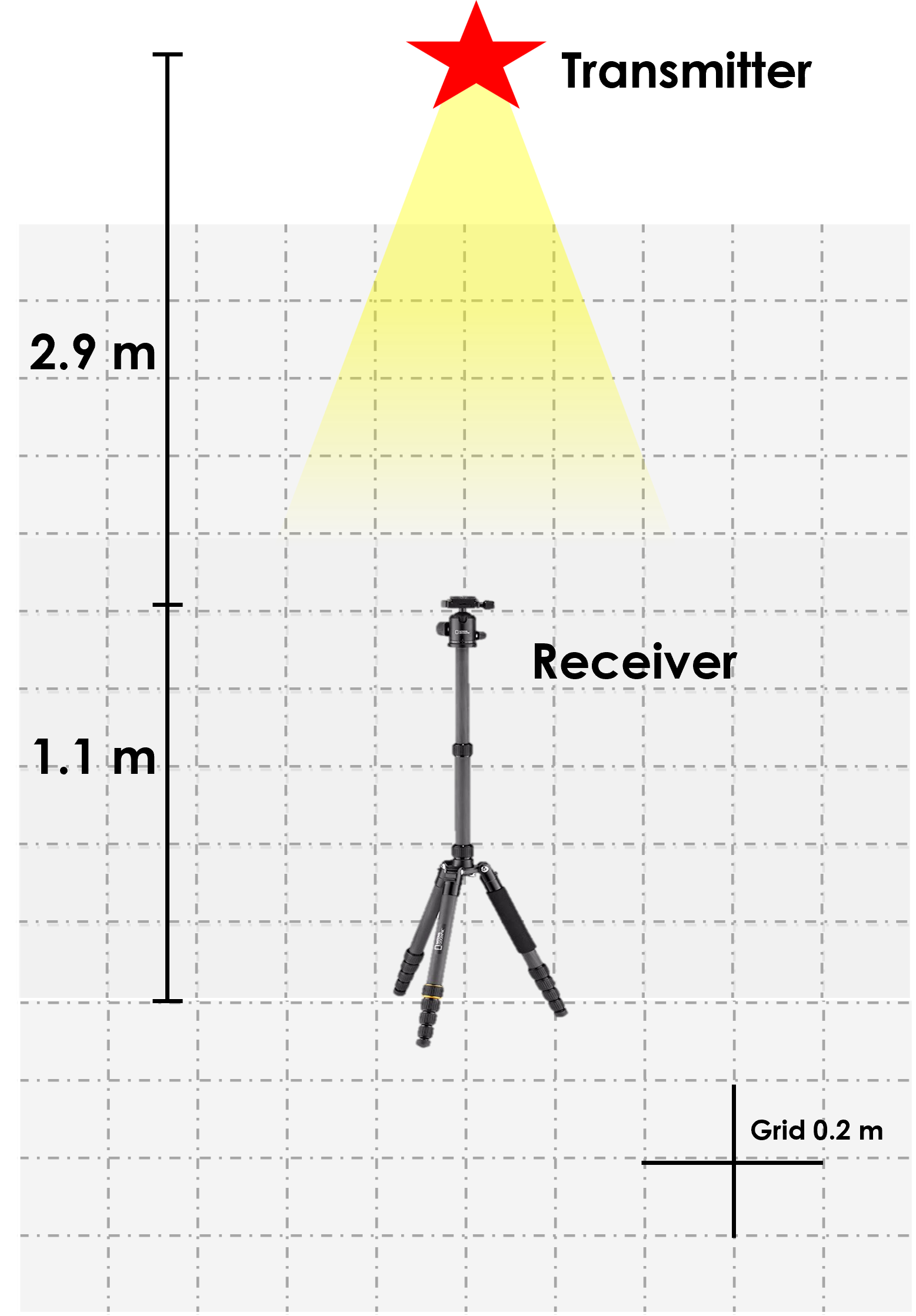}}%
        \label{fig:txrx}%
    }
    \hspace{1.5cm}
    \subfloat[]{%
        \includegraphics[scale=0.65]{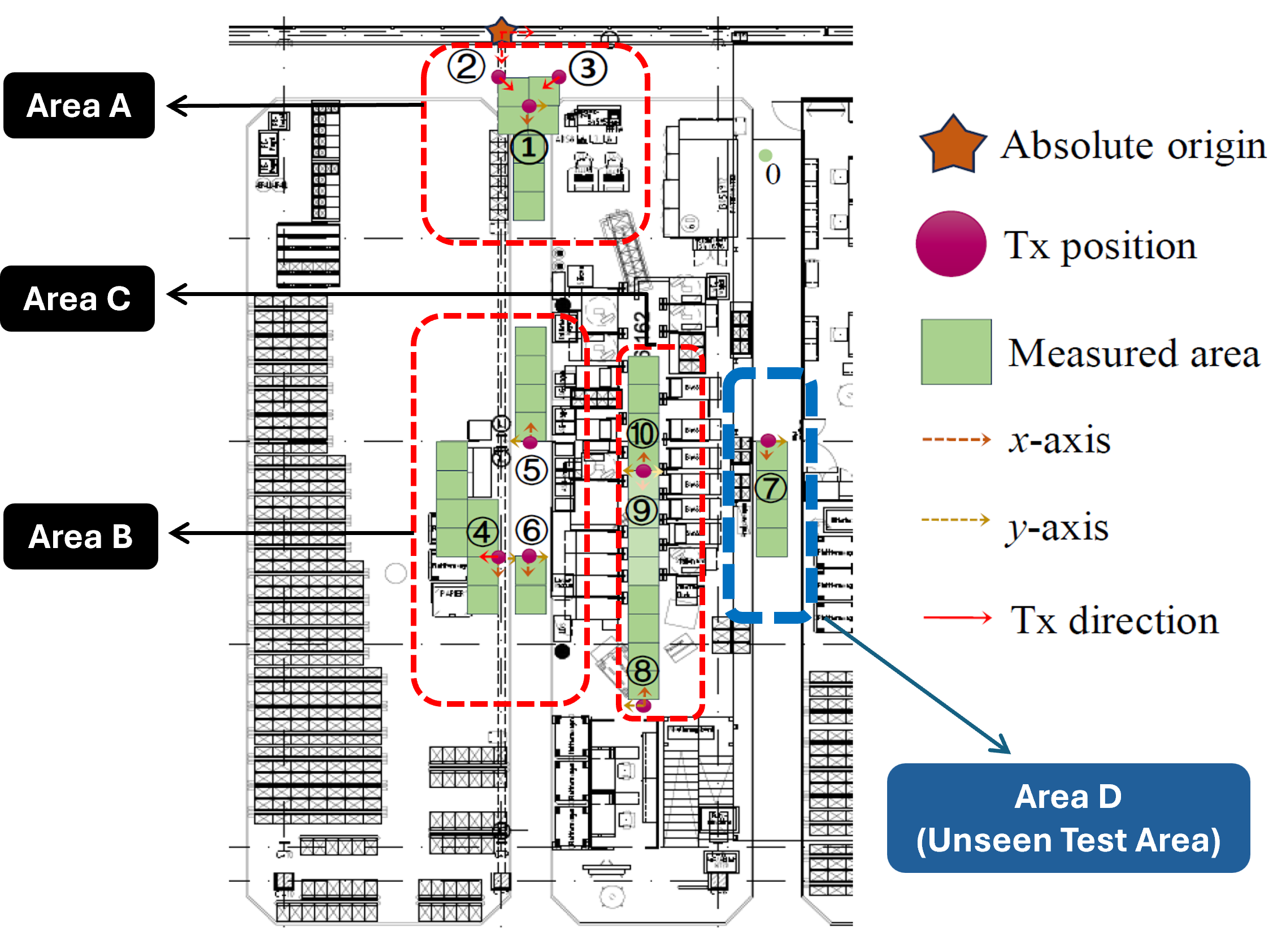}%
        \label{fig:floorplan}%
    }
    \caption{(a) Tx/Rx setup. (b) Floor plan layout of the measurement area.}
\end{figure*}

An overview of the VLC transceiver arrangement and the measurement grid layout is presented in Fig.~\ref{fig:txrx}. To align with the physical constraints and operational setup of the production environment, the Tx and Rx front-ends were mounted at fixed heights of 2.9 meters and 1.1 meters, respectively.
To ensure systematic data collection and facilitate consistent statistical analysis, the receiver area was segmented into a uniform grid with cells measuring 0.2 m $\times$ 0.2 m. Signal measurements were recorded at each grid intersection point, providing a dense and evenly distributed dataset across the coverage area. Fig. \ref{fig:floorplan} displays the actual positions of the Txs and the measured area within the production line. The factory-provided floor plan is annotated with the precise locations of Txs, measurement zones, and the absolute reference origin (marked with an orange star), which serves as the global coordinate starting point for all data acquisition. The measurement campaign encompasses ten distinct Tx deployment areas (pt.~\Circled{1}--\Circled{10}), distributed within Areas A, B, and C. As demarcated on the map by dashed red boundaries, pt.~\Circled{1}--\Circled{3} belong to Area A, pt.~\Circled{4}--\Circled{6} belong to Area B, pt.~\Circled{8}--\Circled{10} corresponds to Area C. Notably, deployment pt.~\Circled{7} outlined in blue—was intentionally excluded from model training and is designated as Area D, used solely to assess the generalization capability of the using the TES framework. The four deployment areas (A–D) were selected to represent structurally and operationally diverse environments, including differences in ceiling structures, lighting conditions, reflective surfaces, and equipment density, providing realistic variations for cross-area TL evaluation.

Each green-highlighted region indicates the coverage zone where RSS measurements were collected. On average, around 160 Rx positions were sampled per area. At each Rx position, the received optical signal was captured continuously for 5 seconds to account for temporal variability induced by ambient noise, equipment movement, and other environmental dynamics. The collected RSSI samples at each point were then averaged to produce a single robust signal strength value, effectively mitigating the impact of short-term disturbances and enabling more stable localization model training. Detailed 3D positions and corresponding area assignments of all Tx deployments are summarized in TABLE~\ref{tab:tx_positions}. 

\begin{table}[!t]
\centering
\caption{3D coordinates, placement context, and deployment areas of transmitters.}
\label{tab:tx_positions}
\renewcommand{\arraystretch}{1.5}
\resizebox{8.5cm}{!}{%
\begin{tabular}{|c|c|p{4cm}|c|}
\hline
\textbf{Tx ID} & \textbf{Position (x, y, z)} & \textbf{Physical Placement Context} & \textbf{Deployment Area} \\
\hline
~\Circled{1} & (2.6, 0.8, 2.9) & Near control station and transparent windows & Area A\\
\hline
~\Circled{2} & (1.7, -0.2, 2.9) & Near control station and transparent windows & Area A\\
\hline
~\Circled{3} & (3.7, 1.8, 2.9) & Near control station and transparent windows & Area A \\
\hline
~\Circled{4} & (17.1, 0.2, 2.9) & Material storage zone & Area B \\
\hline
~\Circled{5} & (12.9, 0.8, 2.9) & Rear of production machinery (beside cement pillars) & Area B\\
\hline
~\Circled{6} & (17.1, 0.8, 2.9) & Rear of production machinery (adjacent to storage) & Area B \\
\hline
~\Circled{7} & (14.2, 8.5, 2.9) & Beneath conveyor belts near lockers & Area D \\
\hline
~\Circled{8} & (23.2, 4.6, 2.9) & Main production line & Area C \\
\hline
~\Circled{9} & (15.2, 4.6, 2.9) & Main production line & Area C \\
\hline
~\Circled{10} & (13.5, 3.6, 2.9) & Main production line & Area C \\
\hline
\end{tabular}}
\end{table}

\section{Performance Evaluation and Analysis} 
\label{sec:results}

This section presents a detailed evaluation of the proposed TL framework for VLC-based localization across different subareas of the Bosch factory hall. We analyze (i) the significance of TL when applying models across domains, (ii) the rationale for selecting the optimal fine-tuning data ratio, and (iii) TES-driven selection of the optimal TL source model and validation on an unseen area.

\subsection{Impact of TL on Cross-Domain Localization}
\label{imp}

To evaluate the necessity of TL, a DNN was trained on Area~A and tested both on its own validation set and on datasets from Areas~B and C. The training used the settings summarized in Table~\ref{tab:TL_simulation_settings}. 

\begin{table}[!t]
\centering
\caption{TL Simulation Settings}
\renewcommand{\arraystretch}{1.3} 
\setlength{\tabcolsep}{6pt} 

\begin{tabular}{@{} l l c @{}} 
\toprule
\textbf{Parameter} & \textbf{Description} & \textbf{Value} \\
\midrule
Optimizer & Model optimizer & \texttt{ADAM} \\
Learning rate, $\eta$ & Step size in gradient descent & \texttt{0.0001} \\
Batch size, $B$ & Number of samples per batch & \texttt{64} \\
Epochs per iteration, $E$ & Number of training epochs & \texttt{1000} \\
\addlinespace
Input dimension, $d$ & Size of input feature vector & \texttt{10} \\
Hidden layers, $L$ & Number of hidden layers & \texttt{5} \\
Hidden layer size, $h$ & Number of neurons per hidden layer & \texttt{512} \\
\bottomrule
\end{tabular}
\label{tab:TL_simulation_settings}
\end{table}

\begin{figure*}[!t]
    \centering
    \begin{minipage}{0.48\textwidth}
        \centering
        \includegraphics[width=\linewidth]{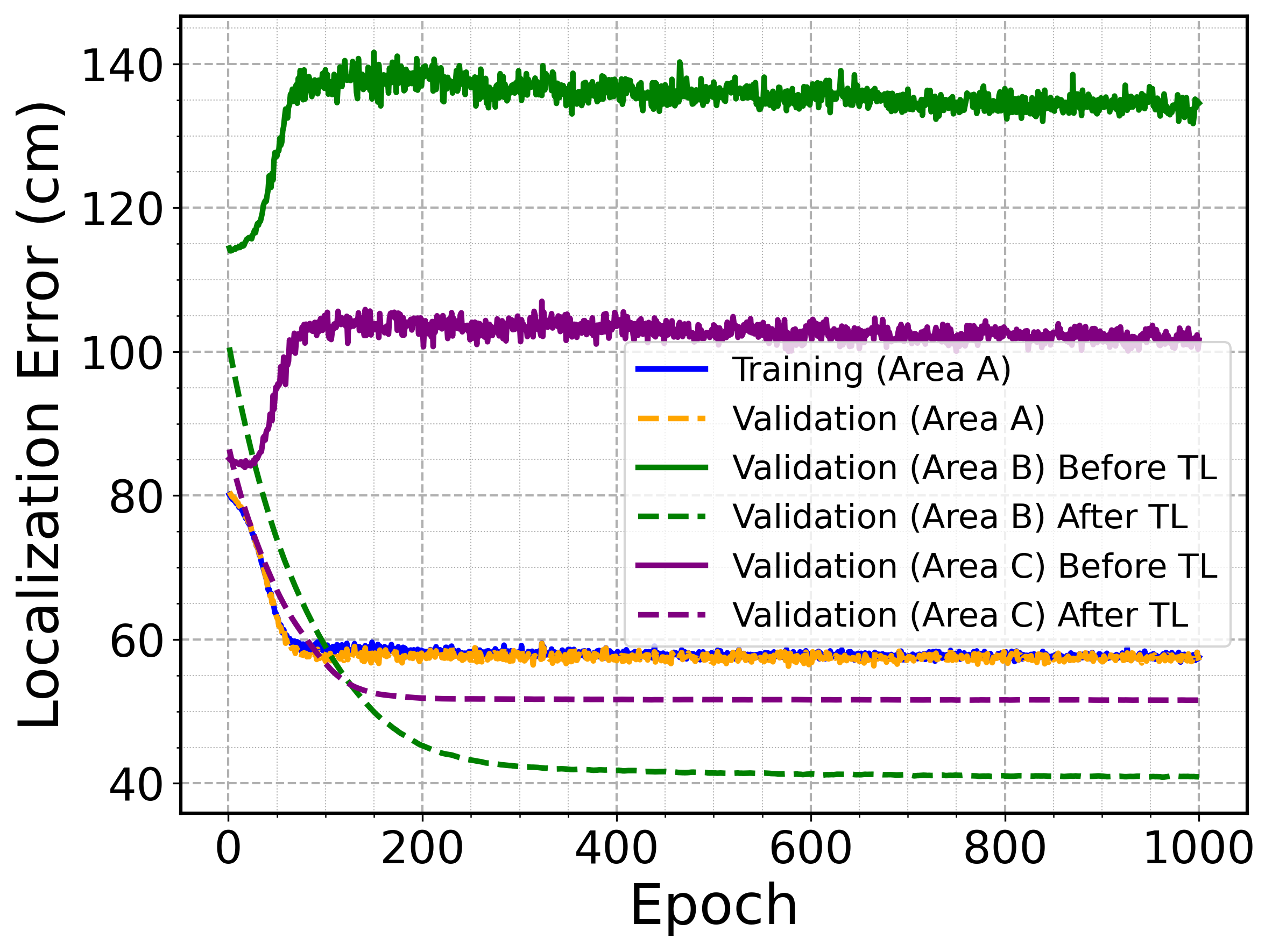}
        \subcaption{}
        \label{fig:loc_error_A}
    \end{minipage}
    \hfill
    \begin{minipage}{0.48\textwidth}
        \centering
        \includegraphics[width=\linewidth]{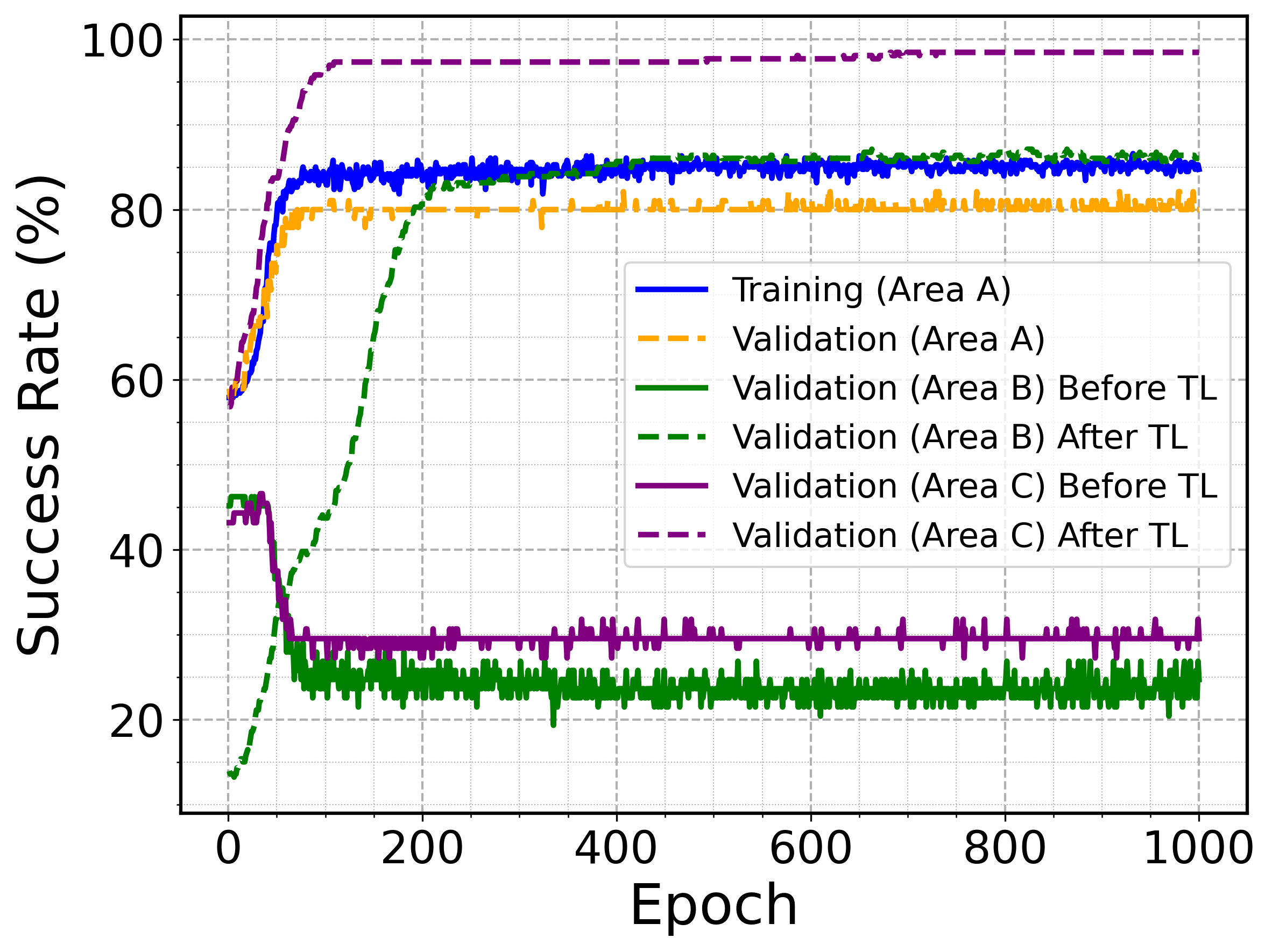}
        \subcaption{}
        \label{fig:success_rate_A}
    \end{minipage}
    
    \vspace{1em} 
    
    \begin{minipage}{0.48\textwidth}
        \centering
        \includegraphics[width=\linewidth]{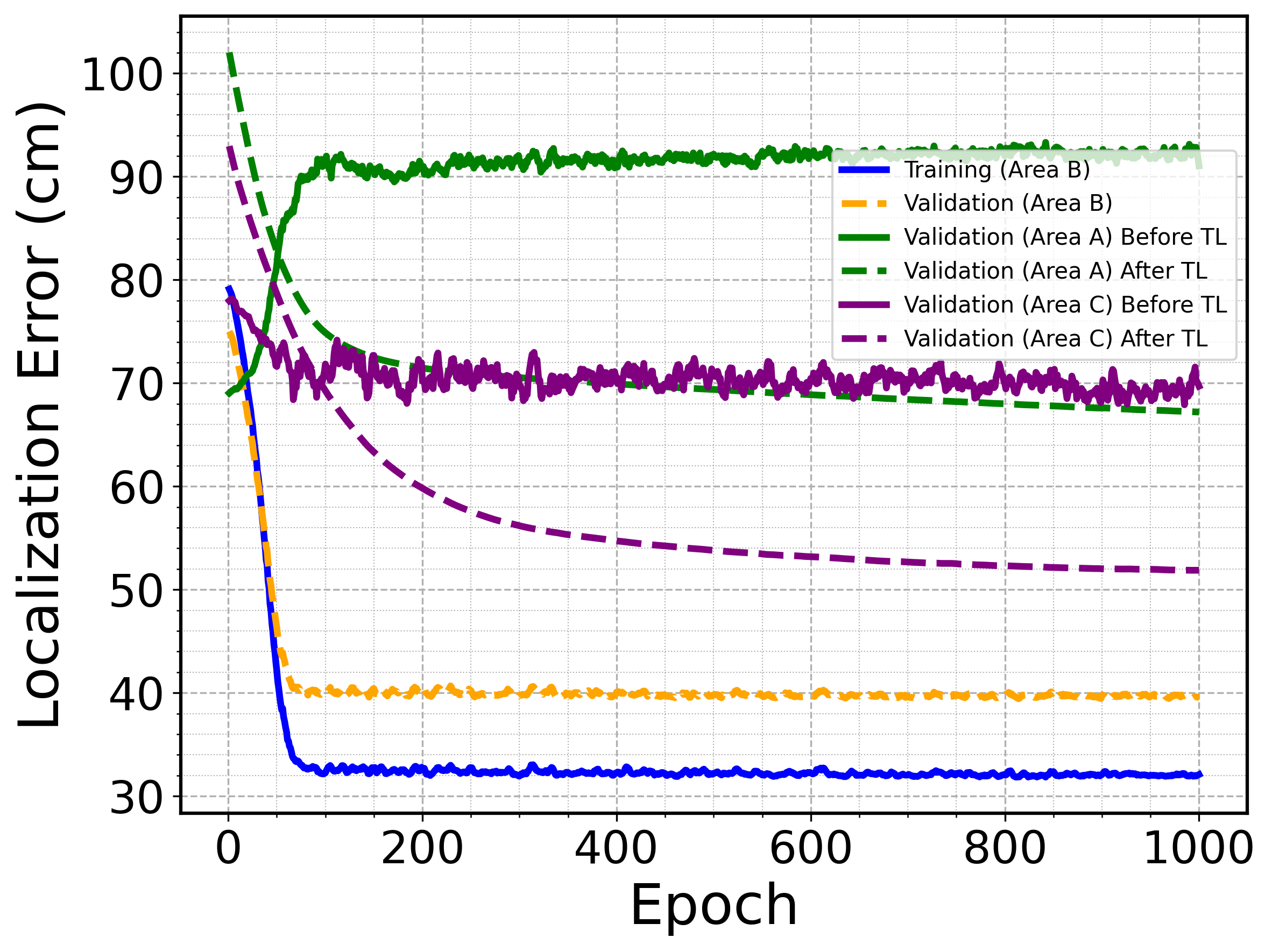}
        \subcaption{}
        \label{fig:loc_error_B}
    \end{minipage}
    \hfill
    \begin{minipage}{0.48\textwidth}
        \centering
        \includegraphics[width=\linewidth]{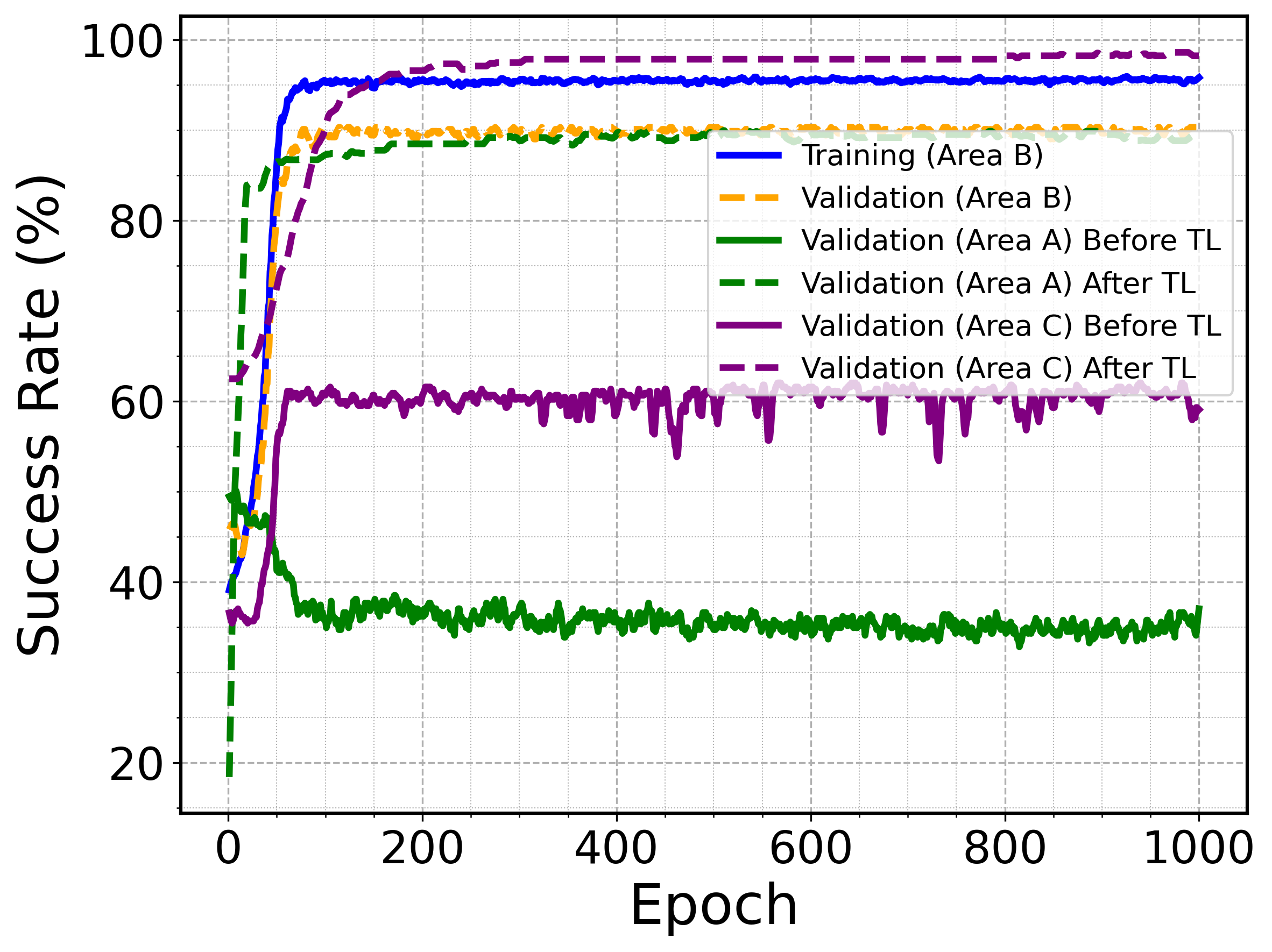}
        \subcaption{}
        \label{fig:success_rate_B}
    \end{minipage}
    
    \vspace{1em} 
    
    \begin{minipage}{0.48\textwidth}
        \centering
        \includegraphics[width=\linewidth]{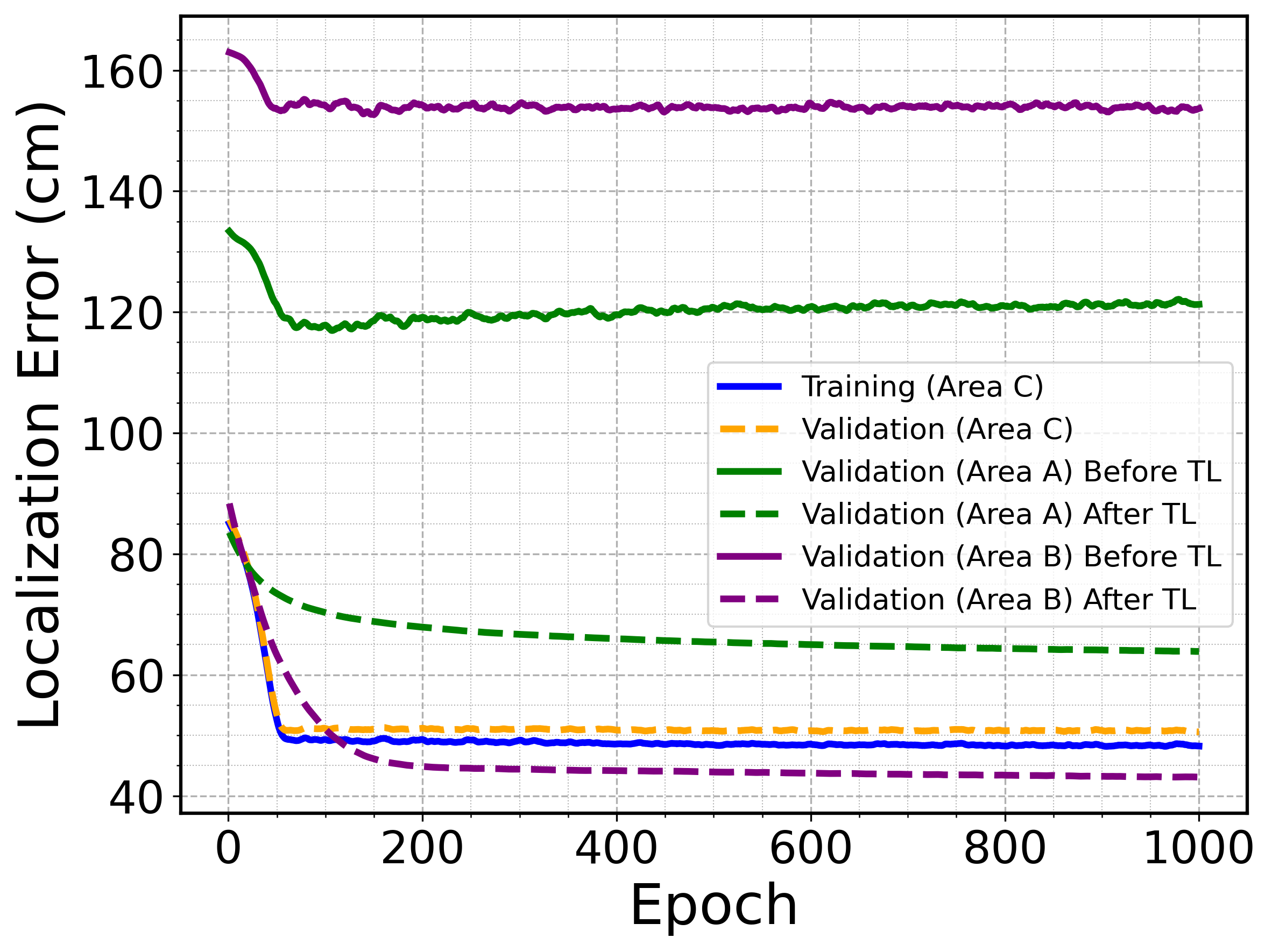}
        \subcaption{}
        \label{fig:loc_error_C}
    \end{minipage}
    \hfill
    \begin{minipage}{0.48\textwidth}
        \centering
        \includegraphics[width=\linewidth]{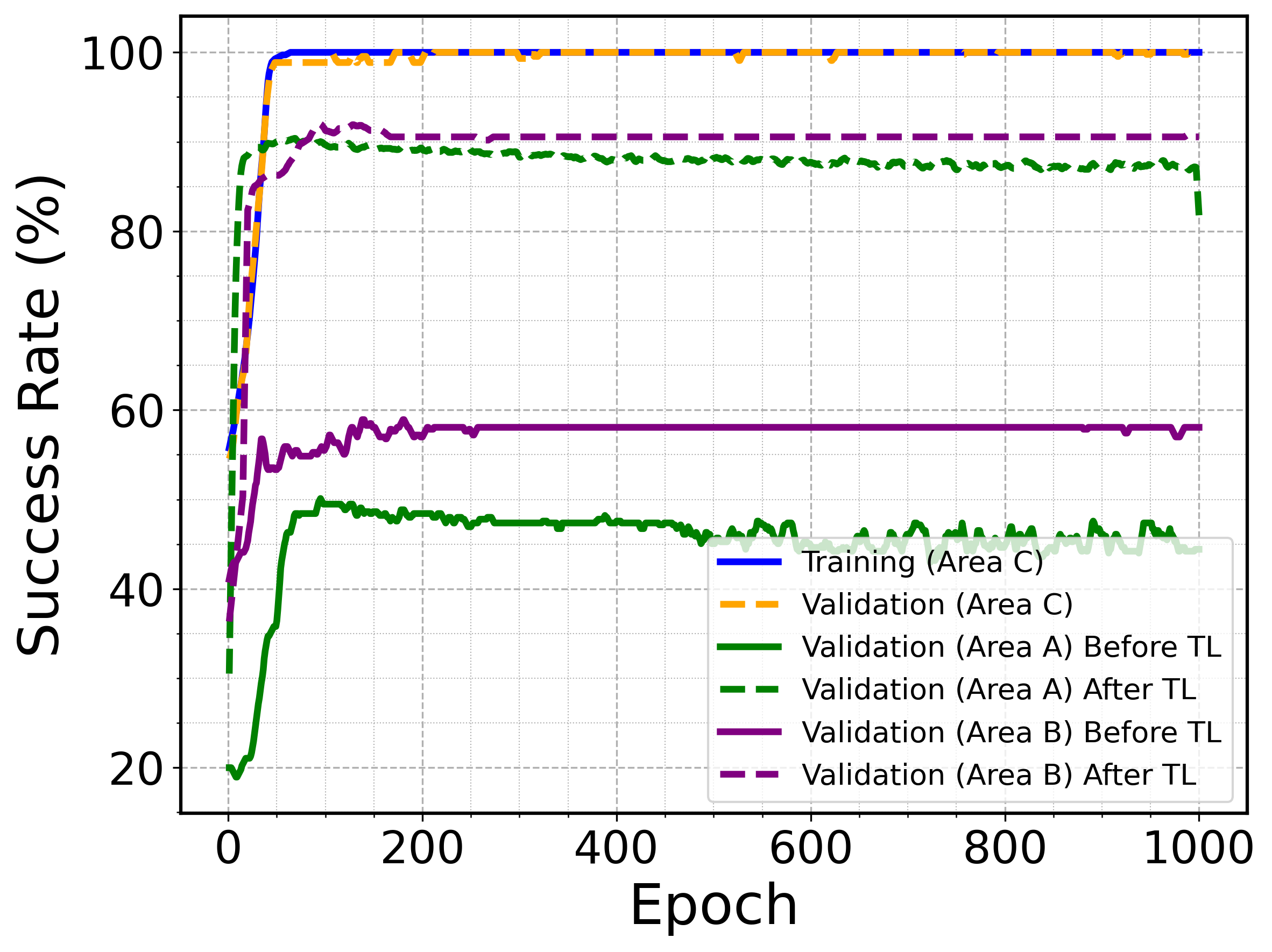}
        \subcaption{}
        \label{fig:success_rate_C}
    \end{minipage}
    
    \caption{TL performance evaluation across different validation areas before and after TL: (a) Area A localization error, (b) Area A success rate, (c) Area B localization error, (d) Area B success rate, (e) Area C localization error, and (f) Area C success rate.}
    \label{fig:performance_eval}
\end{figure*}

Fig.~\ref{fig:performance_eval} shows the validation learning curves for localization error, while Table~\ref{Table:FinalLE} summarizes the final localization errors and success rates after 1000 epochs for all source areas (A, B, and C), evaluated within-domain and cross-domain, with and without TL. On its own validation data, the Area~A model achieved a localization error of 57.73\,cm. Direct application to other areas resulted in substantial performance degradation, with errors rising to 134.91\,cm on Area~B and 101.08\,cm on Area~C.

Applying TL using 40\% of the target area data (chosen based on the analysis in Section~\ref{ft}) restored cross-domain performance. For instance, fine-tuning reduced the Area~A localization error on Area~B to 40.90\,cm, and on Area~C to 51.55\,cm. In some cases, the fine-tuned model even surpassed its original in-domain baseline; for example, the Area~A model fine-tuned on Area~C achieved a success rate of 98.48\%, exceeding its original 80.00\% on Area~A.

These results confirm that models trained on one spatial domain cannot be reliably deployed to another without adaptation. TL not only restores performance but also enhances cross-domain generalization, underscoring its significance for practical VLC-based localization deployments.

\begin{table*}[ht]
\caption{TL results on different areas and metrics. A/B/C in parentheses denote the respective target area.}
    \centering
    \begin{tabular}{llcccccc}
    \toprule
    Source & Metric & Train (Source) & Val. (Source) & \multicolumn{2}{c}{Target 1 (Before/After TL)} & \multicolumn{2}{c}{Target 2 (Before/After TL)} \\
    \cmidrule(lr){5-6} \cmidrule(lr){7-8}
     & & & & Before & After & Before & After \\
    \midrule
    \multirow{2}{*}{Area A} 
        & Error (cm)
            & 57.73 & 57.74
            & 134.91 (B) & 40.90 (B)
            & 101.08 (C) & 51.55 (C) \\
        & Success (\%)
            & 84.47 & 80
            & 23.66 (B) & 86.02 (B)
            & 30.68 (C) & 98.48 (C) \\
    \midrule
    \multirow{2}{*}{Area B}
        & Error (cm)
            & 32.09 & 39.57
            & 91.04 (A) & 67.20 (A)
            & 69.70 (C) & 51.88 (C) \\
        & Success (\%)
            & 95.84 & 90.32
            & 37.05 (A) & 97.19 (A)
            & 59.09 (C) & 98.24 (C) \\
    \midrule
    \multirow{2}{*}{Area C}
        & Error (cm)
            & 48.22 & 50.58
            & 121.28 (A) & 63.88 (A)
            & 153.66 (B) & 43.09 (B) \\
        & Success (\%)
            & 100 & 100
            & 44.42 (A) & 81.78 (A)
            & 58.06 (B) & 90.55 (B) \\
    \bottomrule
    \end{tabular}
    \label{Table:FinalLE}
\end{table*}

\begin{table}[ht]
\centering
\caption{Success rate sensitivity to distance threshold $\delta$ across all source–target transfers.}
\label{tab:SR_thresholds}
\begin{tabular}{lcccc}
\toprule
Transfer & SR @0.5 m & SR @1.0 m & SR @1.5 m & SR @2.0 m \\
\midrule
A$\to$B  & 78\%      & 86\%      & 96\%      & 100\% \\
A$\to$C  & 77\%      & 98\%      & 100\%     & 100\% \\
B$\to$A  & 81\%      & 97\%      & 100\%     & 100\% \\
B$\to$C  & 79\%      & 98\%      & 100\%     & 100\% \\
C$\to$A  & 61\%      & 82\%      & 96\%      & 100\% \\
C$\to$B  & 76\%      & 91\%      & 97\%      & 100\% \\
\midrule
Avg.      & 75\%      & 92\%      & 98\%      & 100\% \\
\bottomrule
\end{tabular}
\end{table}

\textit{Success-Rate Threshold Justification and Sensitivity Analysis:} The localization success rate depends on the selected distance threshold $\delta$ (Eq.~\ref{eq:SR}), which should reflect practical industrial requirements. In large indoor factory environments such as the industrial hall of Bosch considered in this work, a positioning accuracy of about 1 m is generally sufficient for applications such as asset tracking, worker-zone monitoring, and coarse navigation of automated guided vehicles. Given equipment size, aisle spacing, and operational safety margins, $\delta = 1\,\text{m}$ is adopted as a realistic and practically meaningful benchmark throughout the paper.

To assess the dependence of the success rate metric on $\delta$, we evaluated localization performance using thresholds of 0.5 m, 1.0 m, 1.5 m, and 2.0 m across all source–target transfers. The results are summarized in Table~\ref{tab:SR_thresholds}. As expected, the success rate increases monotonically with the threshold, while the relative performance trends between transfer pairs remain consistent. This indicates that TES uses the success rate primarily as a comparative performance indicator rather than an absolute localization-accuracy measure, and that the conclusions of the study are not sensitive to moderate variations in the chosen threshold.

\subsection{Fine-Tuning Data Ratio Justification}
\label{ft}

\begin{figure}[!t]
    \centering
    \begin{minipage}{0.48\linewidth}
        \centering
        \includegraphics[width=\linewidth]{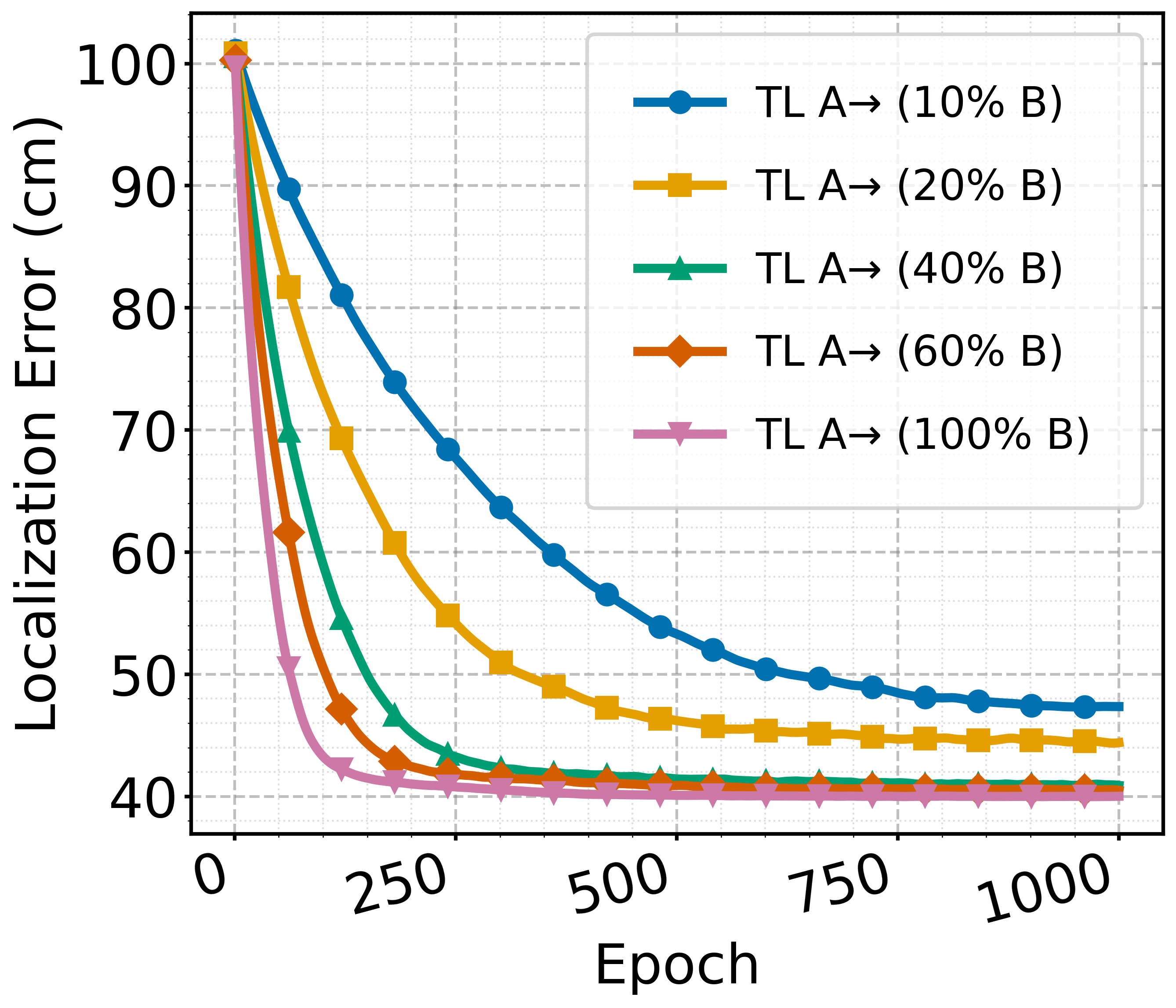}
        \subcaption{}
        \label{fig:loc_error}
    \end{minipage}
    \hfill
    \begin{minipage}{0.48\linewidth}
        \centering
        \includegraphics[width=\linewidth]{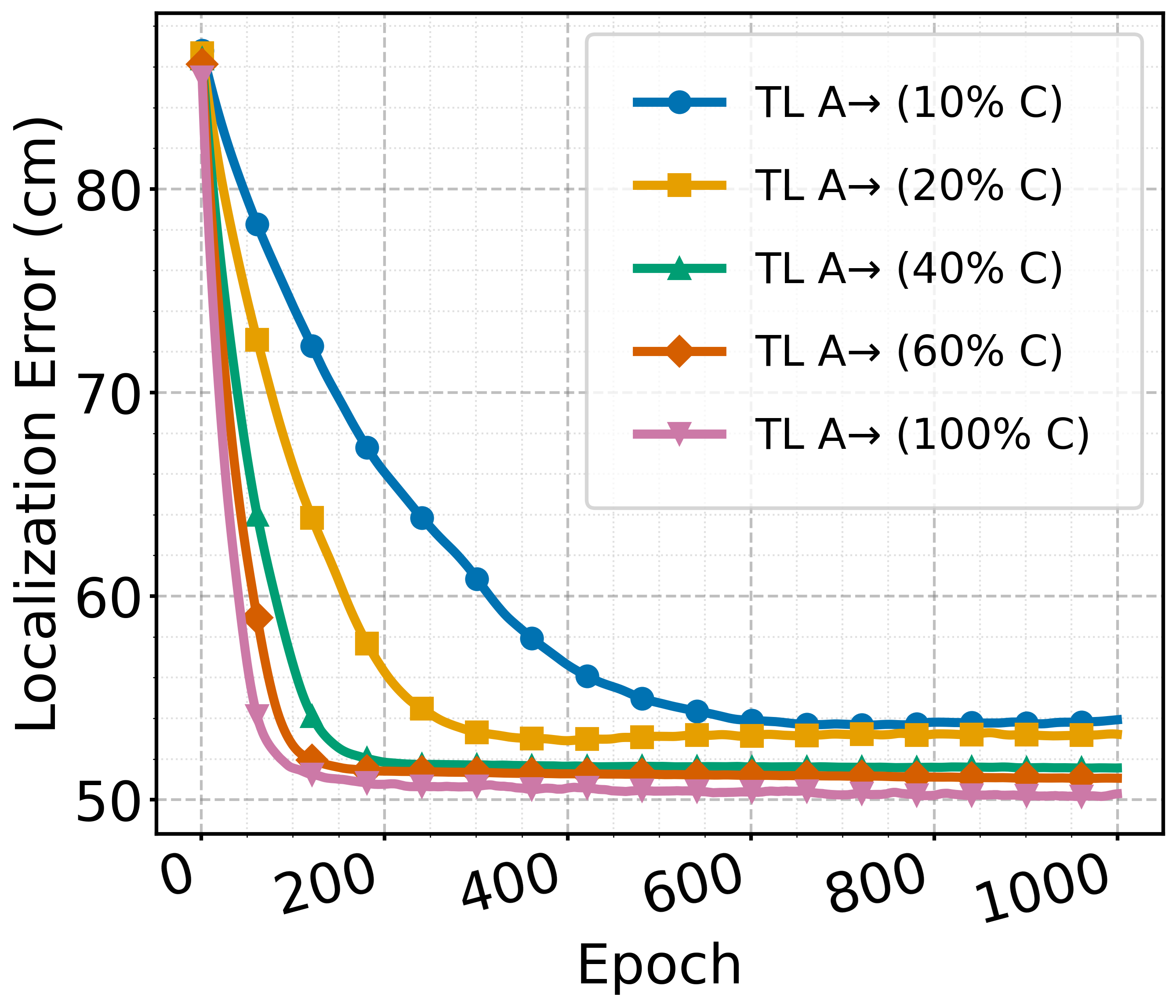}
        \subcaption{}
        \label{fig:success_rate}
    \end{minipage}
    
    \vspace{0.3em} 
    \begin{minipage}{0.48\linewidth}
        \centering
        \includegraphics[width=\linewidth]{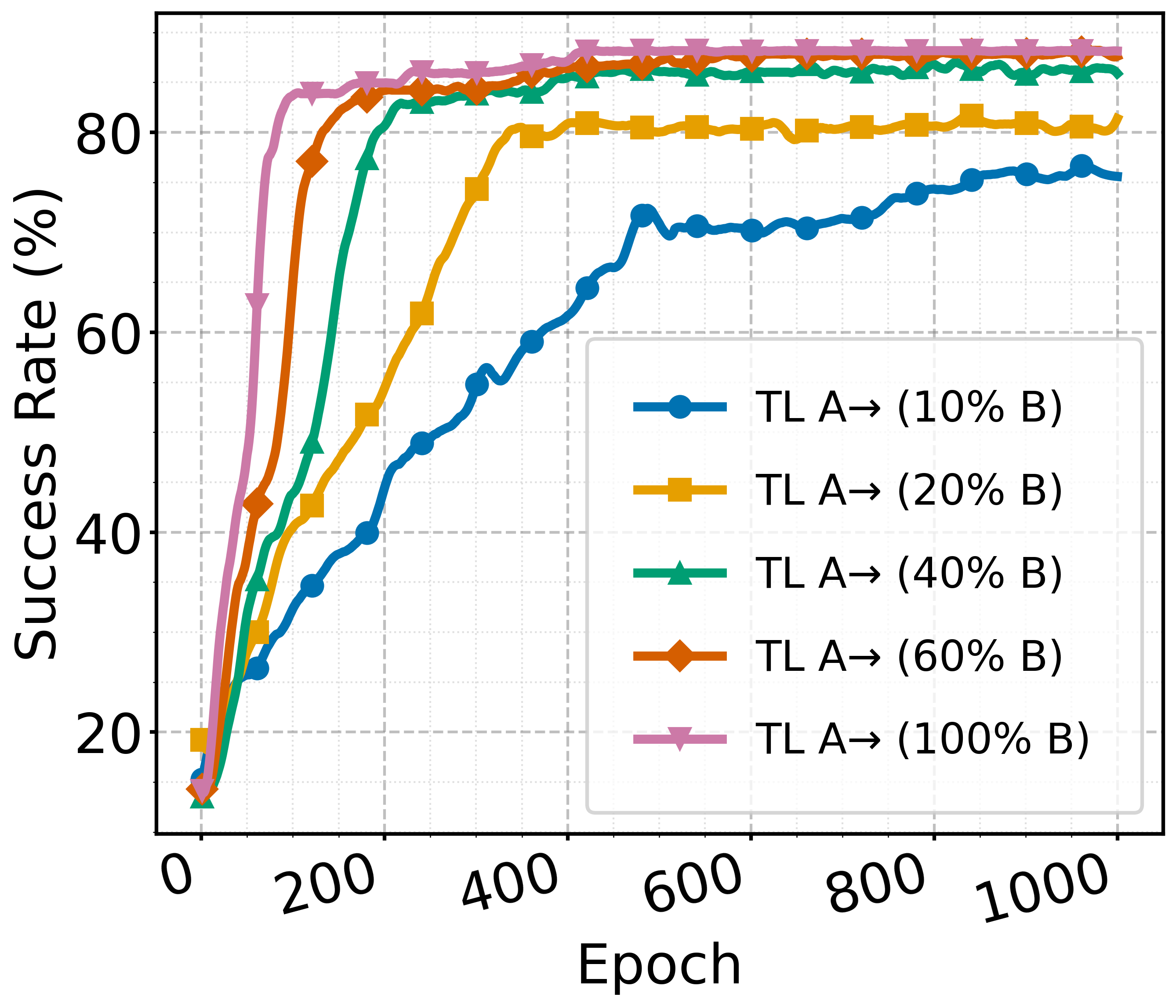}
        \subcaption{}
        \label{fig:image4}
    \end{minipage}
    \hfill
    \begin{minipage}{0.48\linewidth}
        \centering
        \includegraphics[width=\linewidth]{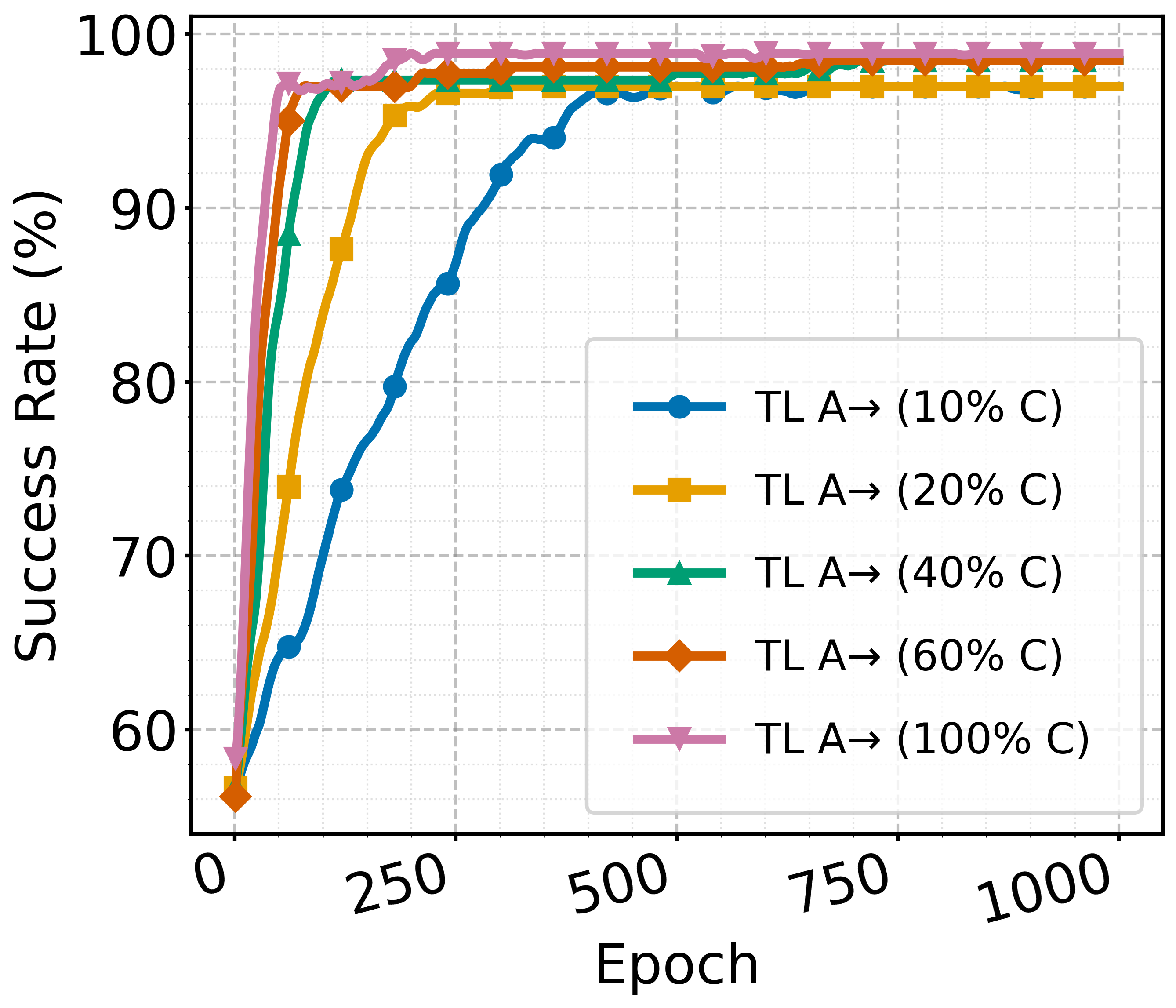}
        \subcaption{}
        \label{fig:image5}
    \end{minipage}

    \caption{Localization error and success rate trends for representative TL scenarios across different fine-tuning data ratios: (a) A$\rightarrow$B localization error, (b) A$\rightarrow$C localization error, (c) A$\rightarrow$B success rate, and (d) A$\rightarrow$C success rate.}
    \label{fig:finetuning_ratio}
\end{figure}

As outlined in Section \ref{imp}, TL was performed using 40\% of the target-domain data. This ratio was determined based on a systematic evaluation of different fine-tuning data proportions. Specifically, we compared TL performance when fine-tuning with 10\%, 20\%, 40\%, 60\%, and 100\% of the available target-area datasets. Fig.~\ref{fig:finetuning_ratio} illustrates the resulting localization errors and success rates.

At low data ratios (10--20\%), TL provides noticeable improvement, but performance remains suboptimal. For example, TL from Area~A to Area~B with only 10\% of the target data yields a localization error of 47.34\,cm and a success rate of 75.63\%. Increasing the ratio to 20\% improves the error to 44.43\,cm and the success rate to 81.36\%. The most significant gain occurs between 20\% and 40\%. At 40\%, the Area~A$\rightarrow$Area~B error drops further to 40.90\,cm with 86.02\% success, while for Area~A$\rightarrow$Area~C, the error decreases to 51.55\,cm with an impressive 98.48\% success rate. Beyond this point, the performance curve flattens: using 60\% or even 100\% of the target data results in only marginal gains, suggesting the model approaches performance saturation. Fig.~\ref{fig:finetuning_ratio} focuses on two representative transfer scenarios (A$\rightarrow$B and A$\rightarrow$C) to avoid clutter, while Table~\ref{finetuning_ratio} reports the complete results for all six source–target domain combinations across all data ratios.

Given these results, we selected 40\% as the standard fine-tuning ratio for all subsequent experiments, as it offers a practical tradeoff between model performance and the data collection effort required in real-world deployments.

The result that approximately 40\% of target-area data recovers most of the full-data performance is an empirical observation for the industrial dataset used in this study. In our earlier work \cite{11174705}, TL was evaluated under lighting variations within the same deployment area, where similar performance recovery was achieved using a comparable calibration fraction. Together, these studies indicate that moderate calibration effort can be effective under both spatial (cross-area) and optical (light/noise) domain variations typical of industrial VLC systems. Nevertheless, the exact calibration ratio may vary with environmental complexity, occlusion, and noise conditions, and should therefore be interpreted as an empirical guideline rather than a universal requirement.

\begin{table*}[ht]
\caption{Localization error (cm) and success rate (\%) for all six source–target transfer scenarios across five fine-tuning data ratios.}
\centering
\begin{tabular}{c|cccccc|cccccc}
\toprule
\multirow{2}{*}{Data \%} &
\multicolumn{6}{c|}{Error (cm)} &
\multicolumn{6}{c}{Success (\%)} \\
 & A$\to$B & A$\to$C & B$\to$A & B$\to$C & C$\to$A & C$\to$B
 & A$\to$B & A$\to$C & B$\to$A & B$\to$C & C$\to$A & C$\to$B \\
\midrule
10  & 47.34 & 53.93 & 71.4  & 58.14 & 68.84 & 46.89  & 75.63 & 96.97 & 93.68 & 97.35 & 87.02 & 92.47 \\
20  & 44.43 & 53.22 & 70.07 & 54.46 & 66.2  & 45.04  & 81.36 & 96.97 & 95.79 & 98.11 & 86.67 & 92.47 \\
40  & 40.90 & 51.55 & 67.2  & 51.88 & 63.87 & 43.09  & 86.02 & 98.48 & 96.84 & 99.11 & 85.96 & 93.55 \\
60  & 40.53 & 51.06 & 66.58 & 51.35 & 63.6  & 42.16  & 87.46 & 98.48 & 97.19 & 99.36 & 87.37 & 93.91 \\
100 & 40.01 & 50.26 & 63.89 & 51.18 & 61.74 & 41.14  & 88.17 & 98.86 & 98.95 & 99.62 & 87.37 & 94.62 \\
\bottomrule
\end{tabular}
\label{finetuning_ratio}
\end{table*}

\subsection{TES-Driven Selection of the Optimal TL Source Model and Validation on an Unseen Area}
\label{subsec:tes-validation}

\begin{figure}[!t]
    \centering
    \begin{subfigure}{0.48\textwidth}
        \centering
        \includegraphics[width=\linewidth]{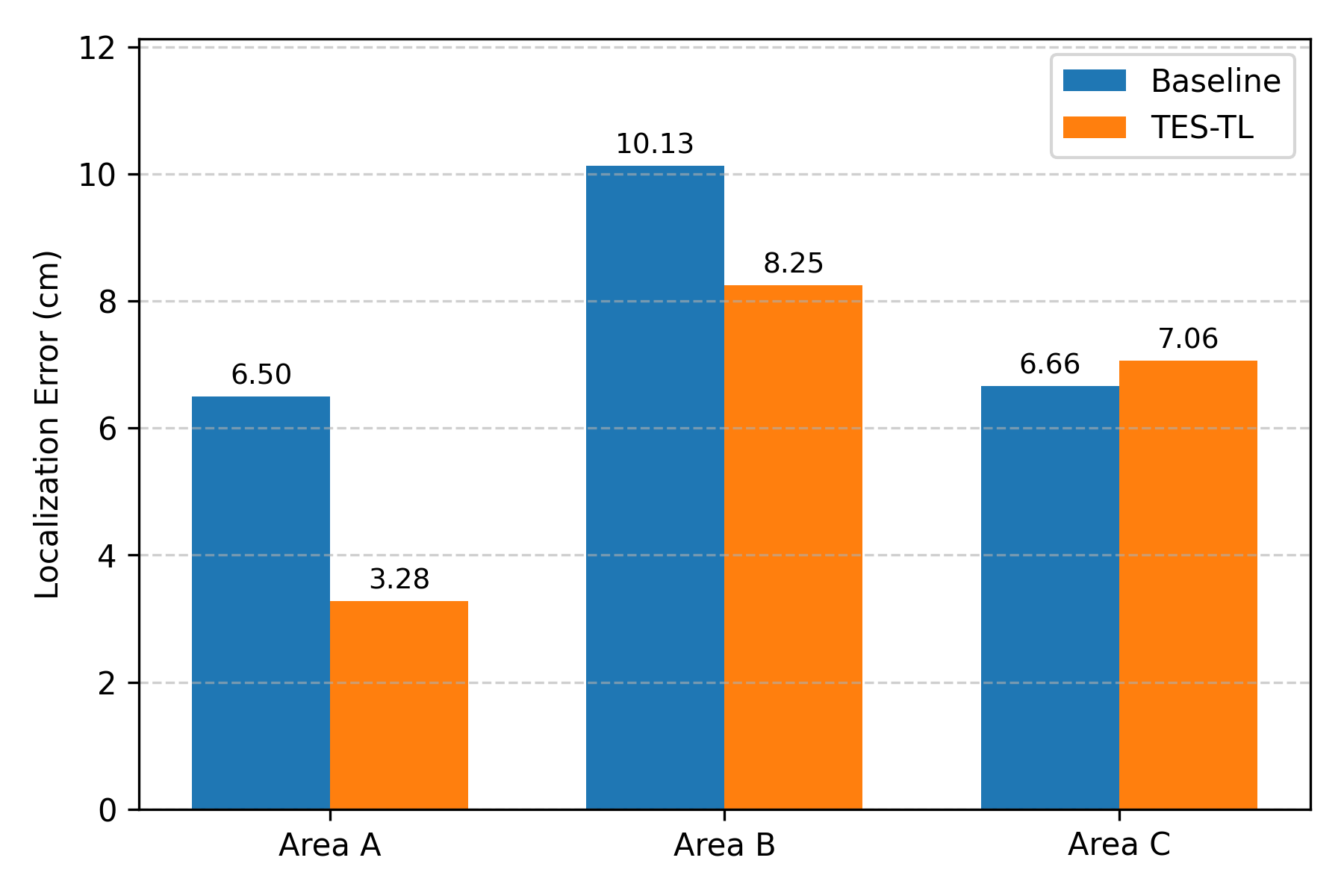}
        \caption{}
        \label{fig:error}
    \end{subfigure}
    \hfill
    \begin{subfigure}{0.48\textwidth}
        \centering
        \includegraphics[width=\linewidth]{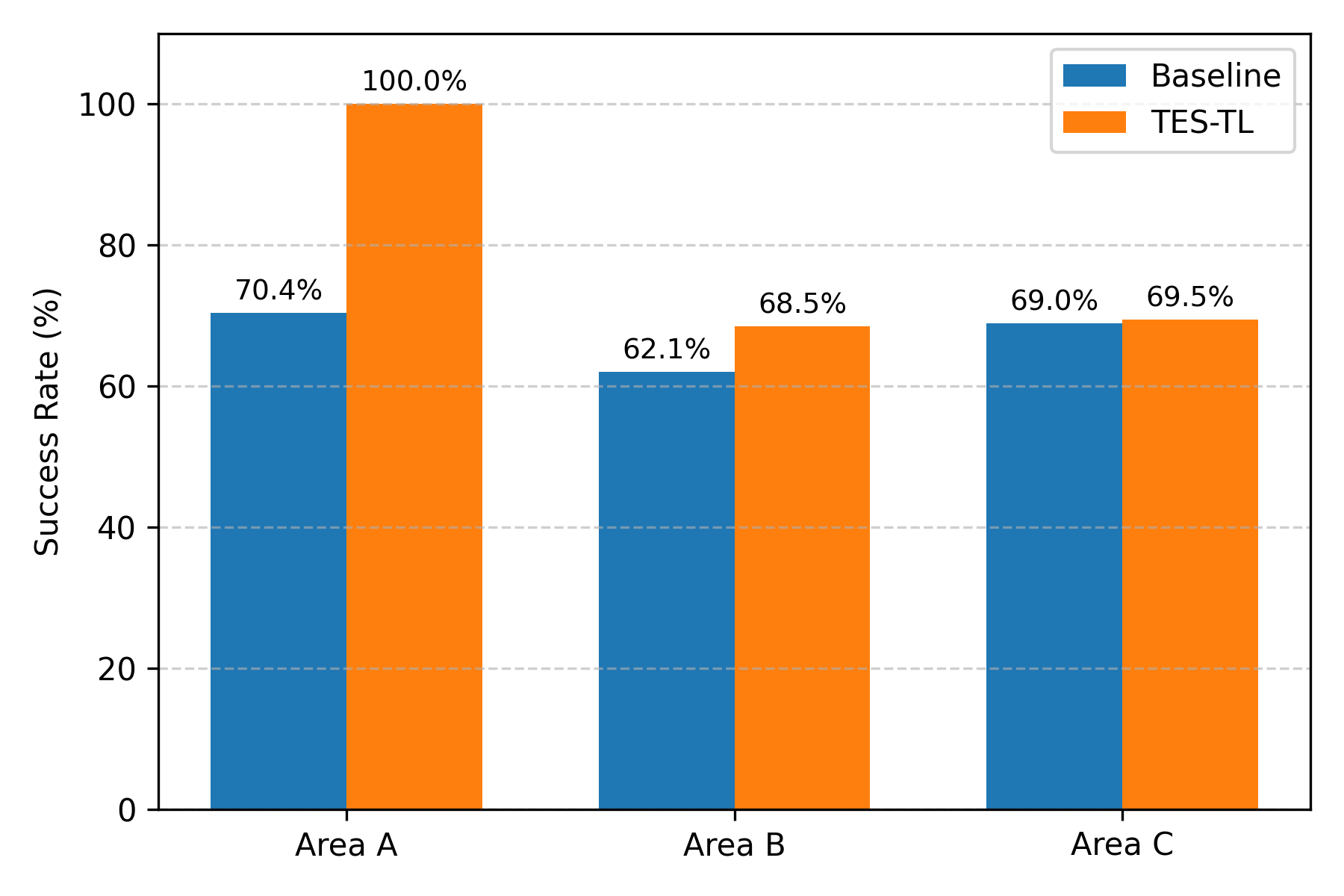}
        \caption{}
        \label{fig:success}
    \end{subfigure}
    \caption{Comparison of TES-selected and baseline models on unseen Area~D: (a) localization error and (b) success rate.}  
    \label{fig:areaD_results}
\end{figure}

\begin{table*}[ht]
\caption{TES results for all source–target pairs, showing post-TL metrics, normalized values, and computed TES.}
\centering
\renewcommand{\arraystretch}{1.2}
\begin{tabular}{l|cc|cc|c|c}
\toprule
\multirow{2}{*}{Source→Target} & \multicolumn{2}{c|}{Post-TL Metrics} & \multicolumn{2}{c|}{Normalized Values} & \multirow{2}{*}{TES} & \multirow{2}{*}{Avg. TES (per source area)} \\
 & Error (cm) & Success (\%) & NE & NSR & & \\
\midrule
A$\to$B & 40.90 & 86.02 & 0.0000 & 0.2537 & 0.6269 & \multirow{2}{*}{\textbf{0.7132}} \\
A$\to$C & 51.55 & 98.48 & 0.4008 & 1.0000 & 0.7996 & \\
\cline{1-7}
B$\to$A & 67.20 & 97.19 & 1.0000 & 0.9215 & 0.4607 & \multirow{2}{*}{0.6224} \\
B$\to$C & 51.88 & 98.24 & 0.4186 & 0.9869 & 0.7841 & \\
\cline{1-7}
C$\to$A & 63.88 & 81.78 & 0.8857 & 0.0000 & 0.0572 & \multirow{2}{*}{0.3876} \\
C$\to$B & 43.09 & 90.55 & 0.0824 & 0.5182 & 0.7179 & \\
\bottomrule
\end{tabular}
\label{tab:TES_combined}
\end{table*}

\begin{table}[ht]
\caption{TES sensitivity to the weighting parameter $\alpha$.}
\centering
\renewcommand{\arraystretch}{1.15}
\begin{tabular}{c|ccc|c}
\toprule
$\alpha$ & Avg TES (A) & Avg TES (B) & Avg TES (C) & Best Source \\
\midrule
0.30 & 0.679 & \textbf{0.755} & 0.336 & B \\
0.40 & 0.696 & 0.688 & 0.362 & A $\approx$ B \\
0.50 & \textbf{0.713} & 0.622 & 0.388 & A \\
0.60 & \textbf{0.731} & 0.556 & 0.414 & A \\
0.70 & \textbf{0.748} & 0.490 & 0.439 & A \\
\bottomrule
\end{tabular}
\label{tab:TES_sensitivity}
\end{table}

To objectively determine which source model offers the most transferable knowledge, we adopt the TES computation pipeline introduced in Section~\ref{sec:sysm} and formulated in Algorithm~\ref{alg:tes_selection_validation} of Section~\ref{sec:algorithm}. Table~\ref{tab:TES_combined} reports post-TL performance metrics, normalized values, and computed TES for all source–target pairs. By averaging TES values for each source’s outgoing transfer paths, the model trained on Area~A achieved the highest mean TES (0.7132), making it the most transferable source model in this study.

\textit{Sensitivity Analysis of TES with respect to the Weighting Parameter $\alpha$:} TES is defined in Eq.~\ref{eq:tes} as a weighted combination of localization error and success rate metrics, where the parameter $\alpha$ controls the balance between accuracy and robustness. In the original evaluation, we adopted $\alpha = 0.5$ to assign equal importance to both components.

To examine the impact of this design choice, a sensitivity analysis was conducted by recomputing TES for $\alpha \in \{0.3, 0.4, 0.5, 0.6, 0.7\}$. For each $\alpha$ value, TES was calculated for every source–target pair (e.g., A$\rightarrow$B, A$\rightarrow$C) using Eq.~\ref{eq:tes}, and then averaged across target areas to obtain the source-level TES ranking, following the same procedure described in Section~\ref{subsubsec:TES}. The detailed TES results for the reference case $\alpha = 0.5$ are reported in Table~\ref{tab:TES_combined}, while Table~\ref{tab:TES_sensitivity} summarizes the corresponding source-level TES values for the different $\alpha$ configurations. The rightmost column of Table~\ref{tab:TES_sensitivity} identifies the source achieving the highest TES, i.e., the most transferable source under each weighting configuration.

As shown in Table~\ref{tab:TES_sensitivity}, the relative ranking of source areas remains largely stable across the tested range. Minor numerical variations appear only at the extremes, reflecting the expected trade-off between robustness-oriented (low $\alpha$) and precision-oriented (high $\alpha$) weighting. Specifically, for low $\alpha$ values, source B slightly outperforms A, whereas for higher $\alpha$, source A gains a clearer advantage. Despite these small shifts, the identification of the most transferable source remains consistent for moderate variations of $\alpha$. These results indicate that the TES-based conclusions are not sensitive to reasonable changes in the weighting parameter, supporting the use of $\alpha = 0.5$ as a balanced operating point.

\textit{Validation of TES using statistical Feature-Space Similarity:} 
To quantitatively examine whether TES reflects physical similarity between deployment areas, we computed the Bhattacharyya distance between the RSS feature distributions of each area pair. Each area is represented by a three-dimensional RSS feature vector corresponding to the three VLC transmitters. For every area, the empirical mean vector and covariance matrix were estimated across all measurement locations, thereby modeling the statistical distribution of the RSS fingerprints within that area.

The Bhattacharyya distance jointly accounts for (i) the Mahalanobis distance between the mean feature vectors and (ii) differences in covariance structure (i.e., variability and correlation patterns). It is computed as
\begin{equation}
\scriptsize
D_B = \frac{1}{8} (\mu_1 - \mu_2)^T \Sigma^{-1} (\mu_1 - \mu_2)
+ \frac{1}{2} \ln \left(
\frac{\det \Sigma}{\sqrt{\det \Sigma_1 \det \Sigma_2}}
\right),
\end{equation}
where $\mu_1$ and $\mu_2$ are the mean RSS feature vectors of the two areas, $\Sigma_1$ and $\Sigma_2$ are their corresponding covariance matrices, and $\Sigma = (\Sigma_1 + \Sigma_2)/2$ denotes the average covariance matrix. The superscript $(\cdot)^T$ denotes the transpose operator, $\det(\cdot)$ denotes the determinant of a matrix, and $\ln(\cdot)$ denotes the natural logarithm. Smaller values of $D_B$ indicate similarity in feature space.

Table~\ref{tab:TES_BD} presents the Bhattacharyya distance and corresponding Bidirectional Avg. TES for all evaluated area pairs. Bidirectional TES is computed as the average of TES values in both directions (source→target and target→source) from Table~\ref{tab:TES_combined}. The results reveal a clear inverse relationship: area pairs with lower Bhattacharyya distance consistently exhibit higher TES, while pairs with larger statistical separation show reduced transfer effectiveness. This confirms that TES is influenced by the underlying similarity of RSS feature distributions and quantitatively validates its sensitivity to physical compatibility between environments.

\begin{table}[ht]
\caption{Bhattacharyya Distance vs. Bidirectional Avg. TES.}
\centering
\setlength{\tabcolsep}{6pt}
\renewcommand{\arraystretch}{1.1}
\resizebox{\columnwidth}{!}{%
\begin{tabular}{c|c|c}
\toprule
Area Pair & Bhattacharyya Distance ($D_B$) & Bidirectional Avg. TES \\
\midrule
A--B & 0.137 & 0.544 \\
A--C & 0.213 & 0.428 \\
B--C & 0.107 & 0.751 \\
\bottomrule
\end{tabular}%
}
\label{tab:TES_BD}
\end{table}

\subsubsection*{Validation on an Unseen Area}

To assess whether TES-based selection generalizes to completely unseen domains, we evaluated all three source models (trained on Areas~A, B, and C) on Area~D (deployment zone P7), which was excluded from all training and fine-tuning. This allowed us to verify not only overall performance but also whether the TES-identified source model (Area~A) indeed provided the best transferability in practice. For each source model, we tested two conditions: \textit{Baseline} (model trained only on its own source area and applied directly to Area~D) and \textit{TES-TL} (the TES-selected source model fine-tuned via TL). Seven independent random 20\% splits of Area~D were used for evaluation, and localization error and success rate were recorded.

As shown in Fig.~\ref{fig:areaD_results}, the TES-TL model originating from Area~A achieved the lowest mean localization error (3.28 cm) and the highest success rate (100\%), outperforming all Baseline models as well as TL models from the other areas. This outcome can be explained by the physical similarity between Areas~A and D. Both share mixed structural characteristics, including reinforced concrete columns and partial boundary openings (windows in A, wide passages and locker space in D), combined with moderate equipment density. These conditions produce comparable signal propagation patterns with limited occlusion and moderate multipath, making knowledge transfer more effective.

By contrast, Area~B (storage/supermarket zone) is relatively open and static, which limited the diversity of its training data and reduced its ability to generalize to Area~D, where large pillars introduce non-trivial shadowing. Area~C, on the other hand, is the busiest production line, characterized by dense machinery and frequent worker movement. Models from C capture strong multipath and temporal variability, which overfit to highly dynamic conditions and transfer less effectively to Area~D’s structurally simpler layout.

Finally, when the unseen-area validation was rotated across all areas (A, B, C, and D), TES consistently selected the source model that shared the closest structural and environmental characteristics with the held-out area. This confirms that TES is not only a statistically reliable metric but also implicitly captures the physical compatibility of environments, making it a robust criterion for identifying the most transferable source model in practical deployment scenarios.

\subsection{Motivation and Impact of Fine-Tuning Strategy}
\label{ftstrategy}

\begin{figure}[!t]
    \centering
    \begin{subfigure}{0.47\textwidth}
        \centering
        \includegraphics[width=\linewidth]{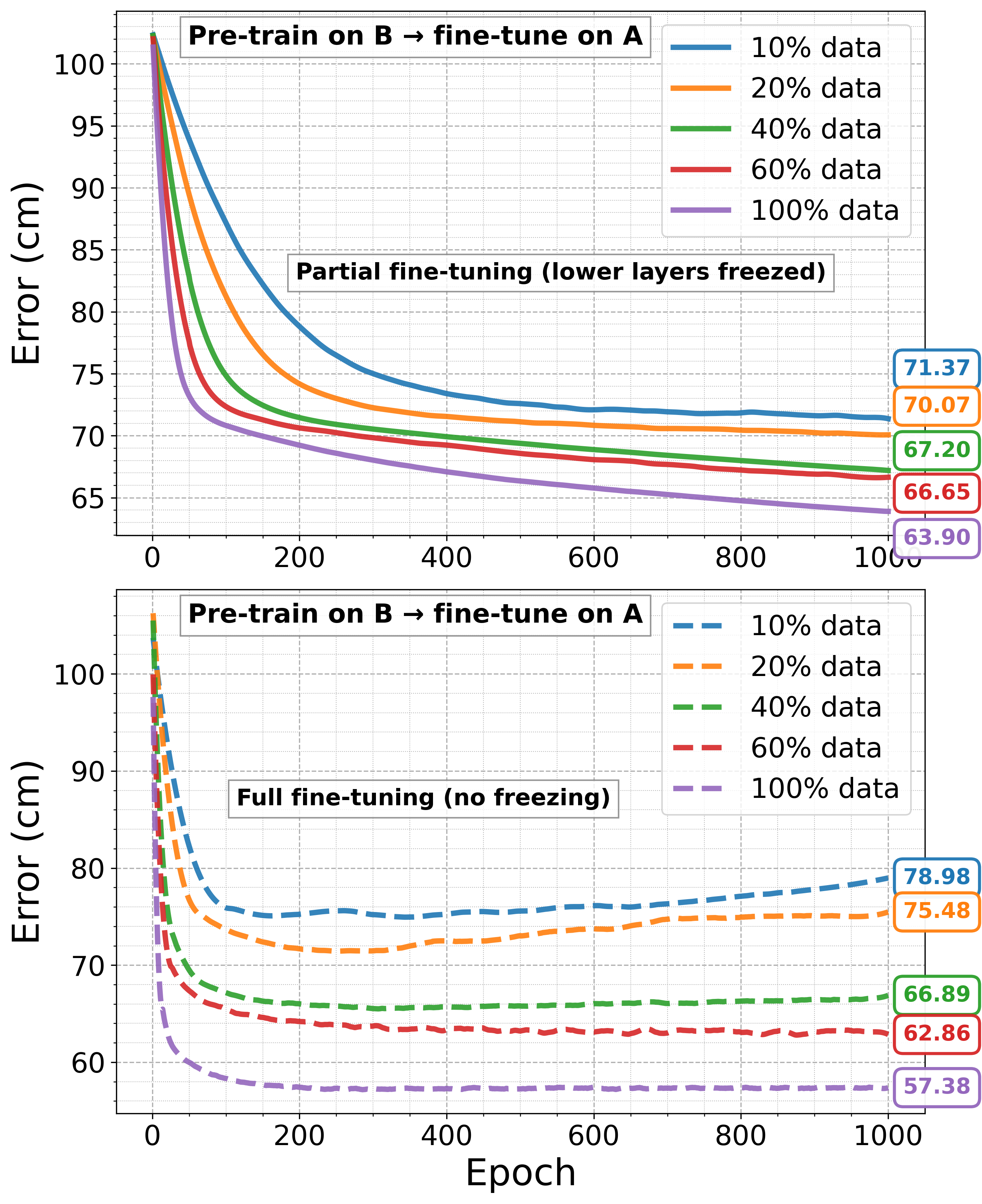}
        \caption{}
        \label{fig:BtoA}
    \end{subfigure}
    \hfill
    \begin{subfigure}{0.47\textwidth}
        \centering
        \includegraphics[width=\linewidth]{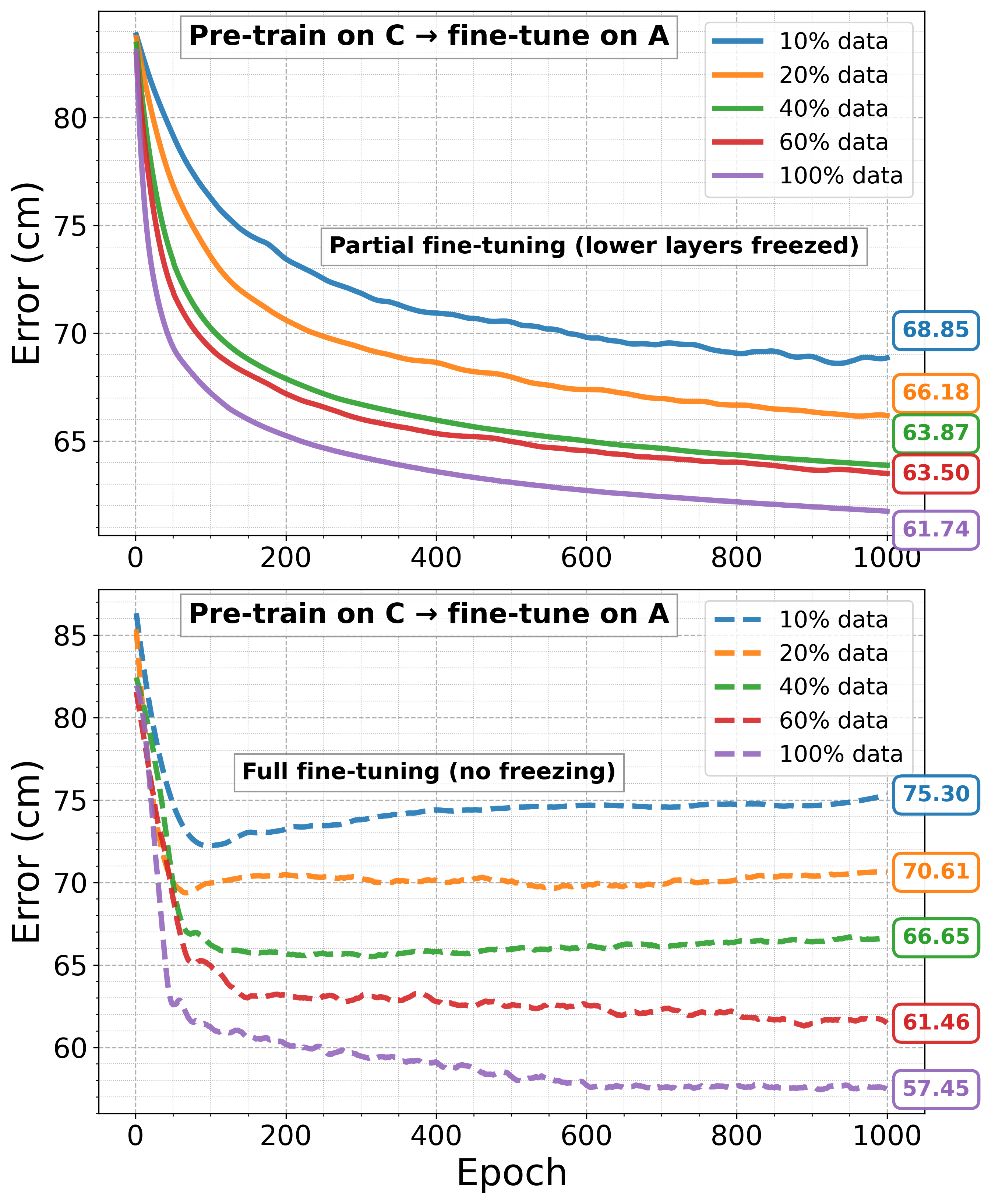}
        \caption{}
        \label{fig:CtoA}
    \end{subfigure}
    \caption{Comparison of partial and full fine-tuning for representative transfers to Area A under different target-data ratios: (a) B$\rightarrow$A transfer and (b) C$\rightarrow$A transfer.}  
    \label{fig:ft_strategy}
\end{figure}

As observed in Tables~\ref{Table:FinalLE} and~\ref{finetuning_ratio}, models trained directly on 100\% of the target-area data can slightly outperform TL models that were pre-trained in a different source area and subsequently fine-tuned using the same complete target dataset. This behavior is consistent with negative-transfer effects reported in deep TL, where source-domain initialization may bias optimization toward parameter regions that are suboptimal for the target distribution. In the present framework, this effect is further influenced by the adopted partial fine-tuning strategy, where lower layers are kept frozen to preserve source-domain representations. This design was motivated by our preliminary observations showing that fully unfreezing the network in low-data regimes can lead to overfitting and unstable training, thereby degrading generalization performance.

To further examine this behavior, we conducted an additional experiment in which only the fine-tuning strategy was modified, while all other factors, including the DNN architecture, optimizer, training protocol, data splits, and source–target transfer settings, were kept unchanged. Specifically, we compared two strategies: (i) partial fine-tuning, where the lower layers remain frozen as used in the proposed data-efficient TL framework, and (ii) full fine-tuning, where all network layers are unfrozen and updated during adaptation.

Fig.~\ref{fig:ft_strategy} shows representative results for the B$\rightarrow$A and C$\rightarrow$A transfer directions. A clear trend can be observed. In low-data regimes, particularly when only 10\%–40\% of the target data is available, partial fine-tuning provides more stable performance by preserving transferable source-domain representations and reducing the risk of overfitting. In contrast, when larger amounts of target data are available, especially in the 60\%–100\% range, full fine-tuning becomes more effective because it allows the layers to adapt all learned representations to the target-area distribution.

For B$\rightarrow$A, partial fine-tuning with 100\% target data reaches 63.90 cm, whereas full fine-tuning reduces the localization error to 57.38~cm. Similarly, for C$\rightarrow$A, the 100\% target-data error decreases from 61.74~cm under partial fine-tuning to 57.45~cm under full fine-tuning. These values are comparable to the target-only baseline of area A reported in Table~\ref{Table:FinalLE}. Therefore, the degradation observed under full target data is not an inherent limitation of TL itself, but rather a consequence of applying a fixed partial-freezing strategy across all data regimes.

These observations further clarify the role of TL within the scope of this work. Under the 100\% target-data setting, fully unfreezing the layers provides sufficient adaptation capacity to better match the target-area distribution and mitigates the degradation observed under partial fine-tuning; however, this behavior should be interpreted as a high-data mitigation strategy, since full fine-tuning in low-data regimes can increase overfitting and training instability. Moreover, acquiring such comprehensive datasets typically requires extensive measurement campaigns, which are often impractical in real industrial settings. Therefore, TL is positioned here primarily as a data-efficiency mechanism, aimed at reducing the need for exhaustive target-area data collection while maintaining high localization performance.

\subsection{Limitations' Discussion and Future Work}
\label{lim}

Although the proposed TL-based VLC localization framework demonstrates effective cross-area adaptation, several practical limitations remain that motivate future work.

(i) \textbf{Single-Facility Evaluation:} The experimental campaign was conducted within a single industrial facility due to practical constraints associated with measurements in active production environments, including safety procedures, coordination with manufacturing schedules, and data confidentiality requirements. While multiple zones with diverse layouts, equipment configurations, and lighting conditions were evaluated, validation across additional industrial facilities remains a limitation. Efforts are underway to extend the VLC dataset with measurements from a second industrial site.

(ii) \textbf{Height Variations:} In the current study, Tx and Rx heights were fixed within the typical installation range of ceiling-mounted LEDs and receiver modules in the industrial facility. In scenarios involving mobile robots or handheld devices, height variations may introduce geometric distortion in RSS patterns and affect localization accuracy. Future work will investigate strategies to address such variations, including augmenting training data with height perturbations, incorporating height as an additional input feature, or jointly estimating position and height within the TL-based localization framework.

(iii) \textbf{Semi-Supervised and Representation-Preserving Adaptation:} The present framework focuses on supervised TL using a limited labeled subset of the target area. This design matches the practical objective of reducing target-area measurement effort while maintaining reliable localization performance. Nevertheless, semi-supervised learning represents a natural extension of the proposed framework, particularly in industrial settings where unlabeled VLC measurements may be easier to collect than accurately labeled position data. Techniques such as pseudo-labeling, consistency regularization, and representation-preserving adaptation could further reduce labeling requirements and improve robustness under larger domain shifts. Such methods may also complement the fine-tuning strategy analyzed in Section~\ref{ftstrategy} by exploiting unlabeled target-domain data while preserving useful source-domain representations. Exploring these semi-supervised extensions is therefore an important direction for future work.

(iv) \textbf{TES Scalability and Deployment Realism:} The proposed TES requires evaluating transfer performance across source–target area pairs using labeled data, which can increase computational cost as the number of deployment areas grows. However, TES is designed as an offline planning metric used during system commissioning phase and therefore does not affect runtime localization performance. In typical industrial VLC deployments, the number of operational zones is limited, and TES evaluation is performed only once when configuring the localization framework, making the associated computational cost manageable.
Nevertheless, we acknowledge that exhaustive TES evaluation may become expensive in very large-scale or rapidly expanding deployments. In such scenarios, TES could be approximated or predicted a priori using unsupervised domain-similarity indicators that do not require labeled target data, such as distributional divergence between RSS feature spaces, feature-alignment metrics, or geometry-aware channel similarity measures derived from VLC propagation characteristics. Such proxy indicators may provide a lightweight mechanism for estimating cross-domain transferability before fine-tuning. Investigating scalable TES approximation strategies is an important direction for future work.

(v) \textbf{Multi-Source or Hybrid Transfer:} The current framework considers transfer from a single source environment at a time. In practical industrial deployments, multiple pre-trained source models may be available. Leveraging these models through hybrid strategies, such as dynamic source selection, ensemble initialization, or representation fusion, could improve localization performance by capturing complementary knowledge from different environments. However, multi-source transfer introduces additional design challenges, including increased computational cost, model selection complexity, and potential negative-transfer interactions between heterogeneous domains. Exploring these trade-offs and developing efficient multi-source adaptation strategies represents an important direction for future work.

(vi) \textbf{Comparison With Advanced Adaptation Techniques:} Several adaptation strategies such as domain-adversarial training and meta-learning have shown promising results in RF-based indoor localization. Although these approaches require additional optimization stages, complex training procedures, or large multi-environment datasets, they represent valuable directions for future research in VLC localization. A systematic comparison between TL and other advanced adaptation methods will be investigated in future work to better understand trade-offs among accuracy, data requirements, computational complexity, and deployment practicality in industrial VLC environments.

\section{Conclusion}
\label{sec:conclusion}
In conclusion, this work introduced a TL framework that makes VLC-based indoor localization practical for Industry 4.0 deployments. By leveraging pretrained models and selectively fine-tuning on only a fraction of target-area data, we cut labeling requirements by 60\% while retaining nearly full performance. The proposed TES eliminates the need for exhaustive testing by reliably identifying the most promising source model, guiding adaptation across domains. Experiments on four areas of a real Bosch manufacturing hall demonstrated that the TES-guided approach achieves localization errors as low as 40.9 cm with success rates exceeding 86\%, and even delivers 3.28 cm mean error when applied to an unseen area. These results confirm that the framework not only restores but also strengthens cross-domain performance, offering a scalable and data-efficient path for VLC localization in evolving industrial environments. Future work will explore the collection of Wi‑Fi and other wireless data from the Bosch hall to investigate TL between sensing technologies, enabling cross‑modal localization.

\section*{Acknowledgment}

\noindent The authors gratefully acknowledge the financial supports of the EU Horizon 2020 program towards the 6G BRAINS project H2020-ICT 101017226.

\bibliographystyle{IEEEtran}
\bibliography{references}


\vspace{10pt}
\begin{IEEEbiography}[{\includegraphics[width=1in,height=1.25in, clip,keepaspectratio]{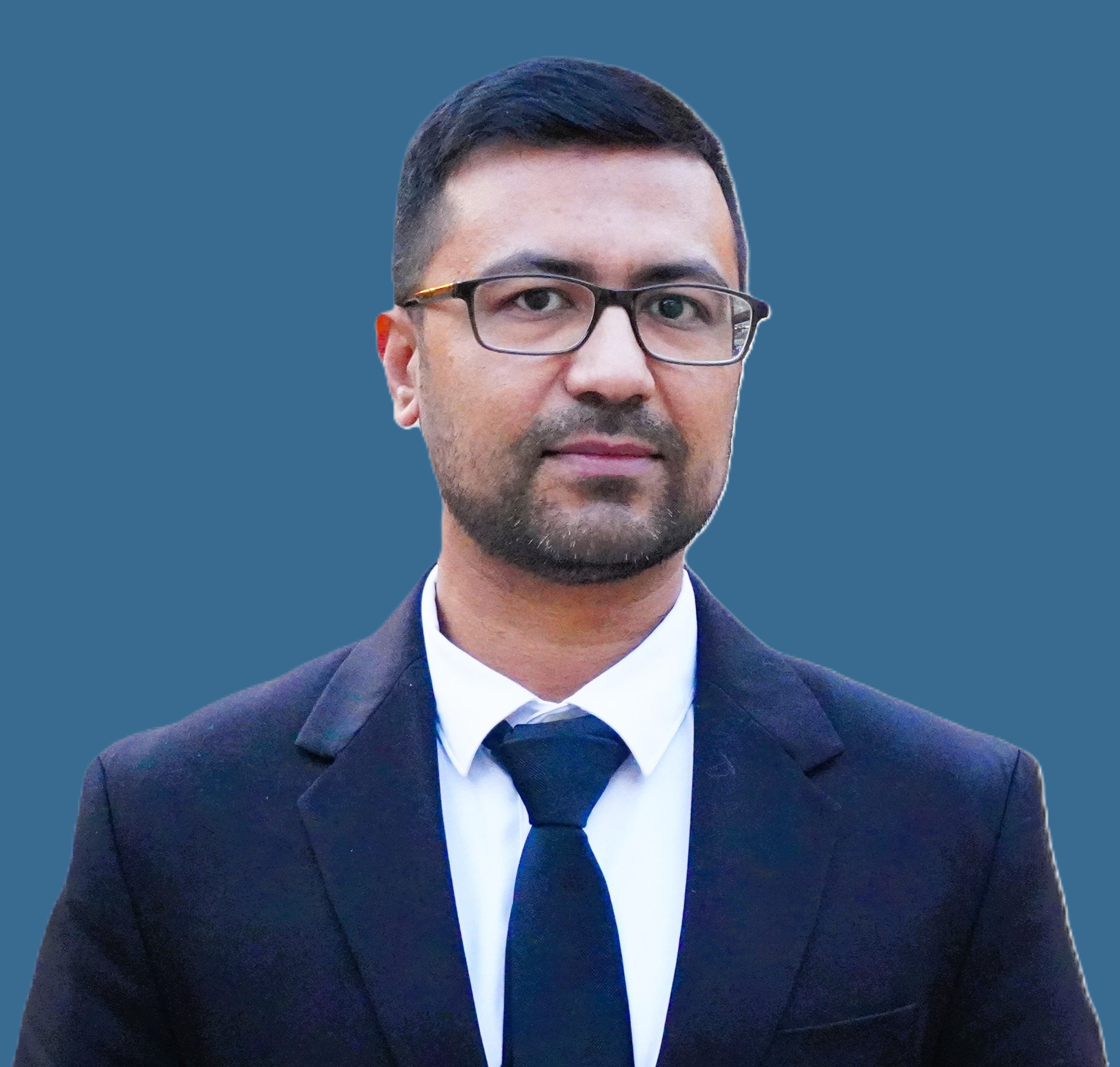}}]{Masood Jan} (Graduate Student Member, IEEE) received the engineering degree in electronics engineering from the University of Engineering and Technology (UET), Peshawar, Pakistan, in 2008, and the master’s degree in telecommunications and networks from the Conservatoire National des Arts et Métiers (CNAM), Paris, France, in 2023. He is currently pursuing the Ph.D. degree in machine learning with Sorbonne University, Paris. He was ranked first in the EDITE 2023 Doctoral Campaign. 

Prior to his doctoral studies, he gained over 12 years of experience in the wireless communications industry, working with multinational companies on AI-driven analytics, automation, and network optimization for LTE/5G systems. His current research interests include machine learning in wireless communication and efficient AI models for 6G/OWC. He is actively engaged in academic dissemination and has delivered numerous talks at conferences and seminars.

\end{IEEEbiography}

\vspace{11pt}
\vfill\break
\begin{IEEEbiography}[{\includegraphics[width=1in,height=1.25in, clip,keepaspectratio]{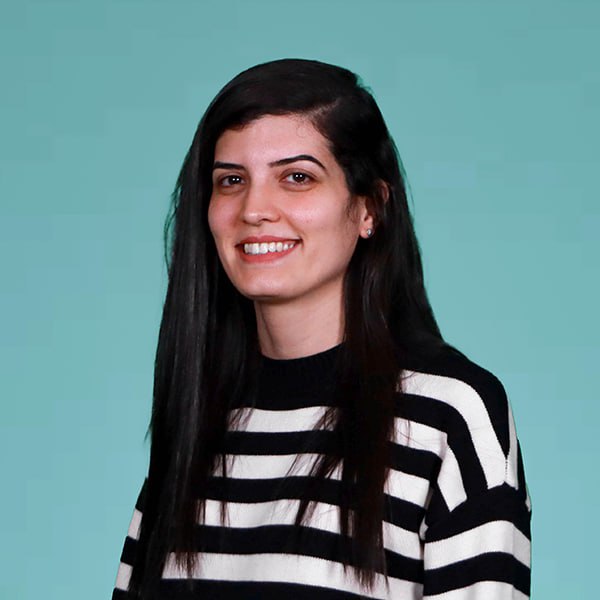}}]{Wafa Njima} (Member, IEEE) received the Eng. degree from the Institut National des Sciences Appliquées et de Technologies de Tunis in 2015. She received the Ph.D. degree in the field of radio-communications in 2019, from the Conservatoire National des Arts
et Métiers in Paris in collaboration with the Ecole Supérieure des Communications de Tunis. She joined the ETIS laboratory at ENSEA Cergy as a temporary assistant and researcher also as a postdoctoral researcher. She is currently an associate
professor at ISEP, Paris. Her publications span several research areas, and her research interests are related to several topics including signal processing, wireless communications, indoor localization, IoT and machine learning for communications. She is IEEE member and she served as TPC member and reviewer for many leading international conferences and journals.
\end{IEEEbiography}

\begin{IEEEbiography}[{\includegraphics[width=1in,height=1.25in, clip,keepaspectratio]{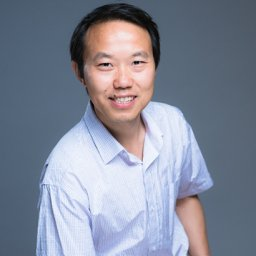}}]{Xun Zhang} (Senior Member, IEEE) is a Professor at ISEP (www.isep.fr), Paris, France, and conducts permanent research at the LISV Laboratory within the University of Saclay-Versailles. As a Senior Member of the IEEE, he serves as an Associate Editor for the IEEE Transactions on Broadcasting and is a member of the BTS Awards Committee. He earned his Ph.D. degree in Instrumentation and Microelectronics (IM) Science from the University of Nancy in October 2009. Between 2009 and 2011, Dr. Zhang undertook a Postdoctoral position at the SCEE Laboratory in CentraleSupélec, France, where he focused on applying auto-reconfigurable management methods and control strategies in FPGAs for cognitive radio systems. In 2011, Dr. Zhang joined the LISITE Laboratory at ISEP as an Associate Professor. His research primarily revolves around PHY layer optimization for 5G cellular networks and Visible Light Communication (VLC) systems. He is also pioneering the development of VLC-based indoor high-accuracy positioning and tracking algorithms.
\end{IEEEbiography}

\begin{IEEEbiography}[{\includegraphics[width=1in,height=1.25in, clip,keepaspectratio]{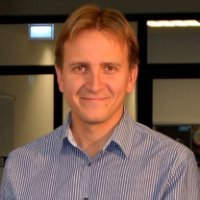}}]{Alexander Artemenko} is a research engineer within the Corporate Sector Research of Robert Bosch GmbH in Stuttgart, Germany. Prior to joining BOSCH in 2015, he was a postdoctoral researcher at the Ilmenau University of Technology (TU Ilmenau), Germany from 2013 to 2015. He received his PhD in engineering from the same university in 2013. In 2012, Dr. Artemenko was a post-doctoral researcher and lecturer at UCLA (Univ. of California Los Angeles) in the group of Professor Mario Gerla.
\end{IEEEbiography}
\vfill

\end{document}